\documentclass{article}

\PassOptionsToPackage{numbers, compress}{natbib}
\usepackage[final]{neurips_2025}

\usepackage[utf8]{inputenc} 
\usepackage[T1]{fontenc}    
\usepackage{hyperref}       
\usepackage{url}            
\usepackage{booktabs}       
\usepackage{amsfonts}       
\usepackage{nicefrac}       
\usepackage{microtype}      
\usepackage{xcolor}         
\usepackage{colortbl}
\usepackage{multirow}
\usepackage{graphicx}         
\usepackage{amsmath} 
\usepackage{pifont}

\title{
  \raisebox{-0.5\height}{\includegraphics[width=2.0em]{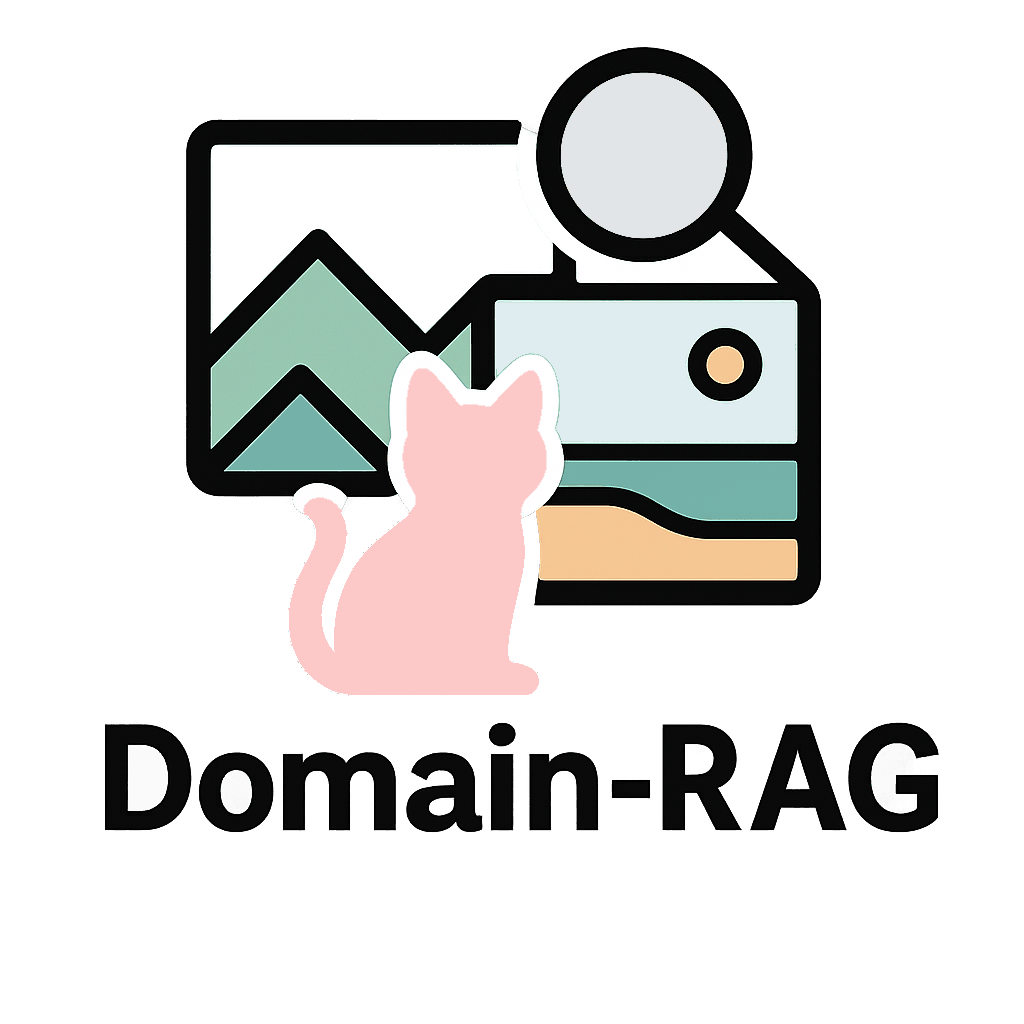}} 
  Domain-RAG: \\ Retrieval-Guided Compositional Image Generation  \\ for Cross-Domain Few-Shot Object Detection}

\author{Yu Li$^{1 *}$ \quad Xingyu Qiu$^{1}$\thanks{These authors have equal contributions.} \quad Yuqian Fu$^{2 *}$\thanks{Corresponding author.} \quad Jie Chen$^{3}$ \quad Tianwen Qian$^{4}$ \quad Xu Zheng$^{2,5}$ \quad \\  \textbf{Danda Pani Paudel$^{2}$} \quad \textbf{Yanwei Fu$^{1}$ \quad  Xuanjing Huang$^{1}$ \quad Luc Van Gool$^{2}$ \quad  Yu-Gang Jiang$^{1}$}
\\
$^{1}$Fudan University \quad $^{2}$INSAIT, Sofia University “St. Kliment Ohridski” \\ $^{3}$Fuzhou University\quad  $^{4}$East China Normal University \quad  $^{5}$HKUST(GZ) 
}

\begin{document}

\maketitle

\begin{abstract}
Cross-Domain Few-Shot Object Detection (CD-FSOD) aims to detect novel objects with only a handful of labeled samples from previously unseen domains. While data augmentation and generative methods have shown promise in few-shot learning, their effectiveness for CD-FSOD remains unclear due to the need for both visual realism and domain alignment. Existing strategies, such as copy-paste augmentation and text-to-image generation, often fail to preserve the correct object category or produce backgrounds coherent with the target domain, making them non-trivial to apply directly to CD-FSOD.
To address these challenges, we propose \textbf{Domain-RAG}, a training-free, retrieval-guided compositional image generation framework tailored for CD-FSOD. Domain-RAG consists of three stages: domain-aware background retrieval, domain-guided background generation, and foreground-background composition. Specifically, the input image is first decomposed into foreground and background regions. We then retrieve semantically and stylistically similar images to guide a generative model in synthesizing a new background, conditioned on both the original and retrieved contexts. Finally, the preserved foreground is composed with the newly generated domain-aligned background to form the generated image. Without requiring any additional supervision or training, Domain-RAG produces high-quality, domain-consistent samples across diverse tasks, including CD-FSOD, remote sensing FSOD, and camouflaged FSOD. Extensive experiments show consistent improvements over strong baselines and establish new state-of-the-art results. The source code and instructions are available at \href{https://github.com/LiYu0524/Domain-RAG}{https://github.com/LiYu0524/Domain-RAG}. 
\end{abstract}

\section{Introduction}
Cross-Domain Few-Shot Object Detection (CD-FSOD)~\cite{fu2024cross}, an emerging task derived from cross-domain few-shot learning (CD-FSL)~\cite{guo2020broader}, aims to tackle few-shot object detection (FSOD) across different domains. 
Unlike conventional FSOD~\cite{kohler2023few}, which assumes source and target data share similar distributions, CD-FSOD considers more realistic scenarios with significant domain shifts, for example, transferring from natural images to industrial anomaly images, remote sensing imagery, or underwater environments. By simultaneously involving the challenges of few-shot learning and domain shift, CD-FSOD poses significant challenges for existing detectors.

Due to the extreme scarcity of labeled data, e.g., as few as 1 or 5 annotated samples per category, a natural and intuitive solution is to leverage data augmentation to alleviate the data bottleneck. Although image augmentation and generation techniques have been extensively studied and shown effective in other few-shot learning tasks~\cite{zhou2023revisiting, li2023knowledge, liang2021boosting, lin2023explore}, it remains unclear whether they can produce high-quality training samples for CD-FSOD. Different from the few-shot classification or in-domain FSOD, this setting requires not only accurate object annotation but also strong domain consistency, which has not been tackled in prior data augmentation-based few-shot learning methods.

\begin{figure}[t]
    \vspace{-0.15in}
    \centering
    \centerline{\includegraphics[width=1.\columnwidth]{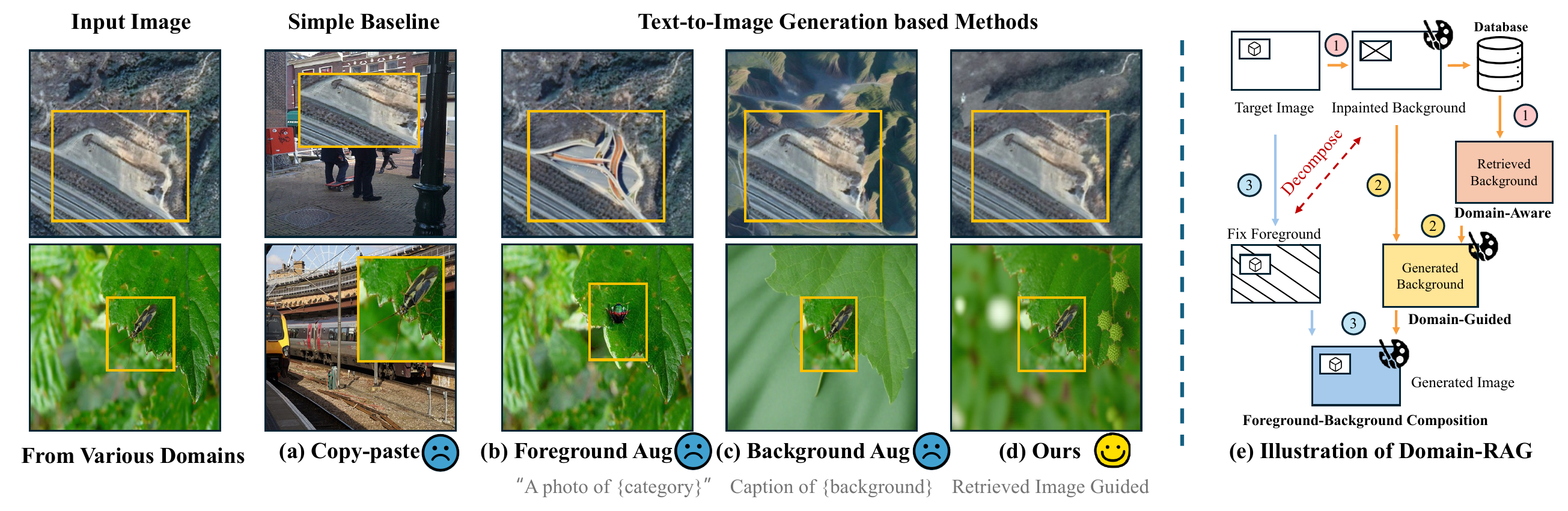}}
    \vspace{-0.1in}
    \caption{ 
    Given images from distinct novel domains, we compare generation results of baseline methods (a–c) and our approach (d), and illustrate the main pipeline of our Domain-RAG (e).}
    \vspace{-0.25in}
    \label{fig:teaser}
\end{figure}

To generate training images for CD-FSOD, the most straightforward approach is copy-paste (Fig.~\ref{fig:teaser}(a)). While easy to implement, such images often lack realism and domain coherence. A more advanced strategy is to build on recent generative models, particularly the trending text-to-image generation, such as SDXL~\cite{podell2023sdxl}, FLUX~\cite{FLUX}. Most existing methods in this area focus on synthesizing foreground objects via text prompts, such as "a photo of {category}" (Fig.~\ref{fig:teaser}(b)). However, they might fail to preserve the object semantics when applied to novel categories and domains. Such a category shift is problematic for CD-FSOD, which has to tackle fine-grained objects and the domain gap. Other approaches generate diverse backgrounds (Fig.~\ref{fig:teaser}(c)), guided by text descriptions of the image. While this better preserves the foreground, purely textual descriptions often fall short of capturing precise domain characteristics and struggle to ensure semantic and visual consistency between foreground and background.  These limitations motivate us to develop a new image generation framework capable of synthesizing visually coherent, domain-aligned training samples for CD-FSOD. Specifically, we aim to: \ding{172} preserve the original foreground object,  \ding{173} generate diverse backgrounds that are both semantically and stylistically aligned with the query image and its domain, and  \ding{174} produce visually realistic images with valid annotations suitable for downstream detection training.

To that end, we propose \textbf{Domain-RAG}, a \textit{retrieval-guided compositional image generation framework} built upon the principle of \textit{fix the foreground, adapt the background}. Leveraging the nature of object detection, Domain-RAG begins by decomposing the target image into its foreground object and background, where the background is recovered by applying an inpainting model~\cite{suvorov2022resolution} to the object-masked region. Although simple in principle, this step is critical for preserving the original object and its annotations, laying the foundation for controllable compositional generation. 
The core challenge then lies in generating a new background that is semantically and stylistically compatible with the foreground. Inspired by the paradigm of retrieval-augmented generation (RAG)~\cite{lewis2020retrieval}, we inject structured visual priors into the generative process to guide background synthesis. As illustrated in Fig.~\ref{fig:teaser}(e), Domain-RAG consists of the following three stages:  
\textbf{1) Domain-Aware Background Retrieval.} We introduce an image database (e.g., COCO~\cite{lin2014microsoft}) containing diverse natural scenes, from which we retrieve candidate backgrounds that are semantically and stylistically similar to the inpainted background of the target image. Semantic similarity is computed using high-level visual features, while style similarity is measured via style-based descriptors~\cite{huang2017arbitrary}.  
\textbf{2) Domain-Guided Background Generation.} Rather than using the retrieved backgrounds directly, we feed them along with the target's inpainted background into a generative model to synthesize a new background that better reflects the visual characteristics of the target domain. To ensure compatibility with modern diffusion models, Redux~\cite{fluxredux2024} is applied to convert visual image cues into descriptive text prompts, enabling direct use of text-to-image generation models.  
\textbf{3) Foreground-Background Composition.} Finally, the preserved foreground is seamlessly composed onto the synthesized, domain-aligned background using a mask-guided generative model. The resulting image maintains the original object while embedding it in a realistic, domain-consistent context (Fig.~\ref{fig:teaser}(d)).
The entire Domain-RAG pipeline is \textit{training-free} and can be directly integrated with existing detectors without any additional supervision or retraining, making it particularly suitable for low-shot scenarios such as 1-shot CD-FSOD.

We validate Domain-RAG on three various tasks that address few-shot object detection with domain shifts: CD-FSOD, remote sensing FSOD (RS-FSOD), and camouflaged FSOD. In all tasks, our method consistently improves a strong baseline by an average of +7.3, +1.1, and +2.1 mAP under the lowest-shot setting, achieving new state-of-the-art (SOTA) performance. These results demonstrate its broad applicability and effectiveness across diverse domains.

Our main contributions are as follows:  
1) We propose Domain-RAG, a training-free, model-agnostic, retrieval-guided compositional image generation framework for boosting cross-domain few-shot object detection.  
2) Domain-RAG enables image generation that preserves the original foreground while synthesizing domain-aligned backgrounds, guided by semantically and stylistically similar retrieved examples. 
3) We achieve consistent performance improvements and new state-of-the-art results across a broad range of CD-FSOD, remote sensing FSOD, and camouflaged FSOD tasks.

\section{Related Works}
\noindent\textbf{Cross-Domain Few-Shot Tasks}.
Few-shot learning across domains has been widely studied~\cite{guo2020broader, tseng2020cross, fu2021meta, liang2021boosting, fu2022wave, zhuo2022tgdm, fu2022me, zhang2022free, fu2023styleadv, luo2023closer, zhuo2024prompt, zou2024closer, peng2024advancing, zhuo2024unified}, but most works focus only on classification. The more realistic task of cross-domain few-shot object detection (CD-FSOD)~\cite{fu2024cross, fu2025ntire}, which involves both recognizing and localizing objects, remains underexplored.
Recent methods like CD-ViTO~\cite{fu2024cross} and ETS~\cite{pan2025enhance} address CD-FSOD. CD-ViTO introduces the task with a closed-source setting (COCO as the only source), while ETS uses a more practical open-source setting~\cite{fu2025ntire} and leverages data augmentation via pretrained GroundingDINO~\cite{liu2024grounding}.
In this paper, we adopt the open-source setting and further improve augmentation using retrieval-guided compositional generation.

\noindent\textbf{FSOD Beyond Domains}.
Beyond classic CD-FSOD tasks, many FSOD or detection problems also involve domain shifts~\cite{cao2024chasing, lu2025musia, lu2025enhancing, liu2024multi, an2025generalized, liu2025ot, anmultimodality, pan2025locate}, even if not explicitly labeled as cross-domain. Two notable examples are Remote Sensing FSOD (RS-FSOD)~\cite{liu2024few} and Camouflaged FSOD~\cite{nguyen2024art}. RS-FSOD uses remote sensing images, which differ from natural scenes in color, perspective, and resolution, creating a clear domain gap. Camouflaged FSOD involves detecting objects that blend into their backgrounds—like fish underwater or animals in the wild—posing challenges for generalization. We include both tasks to assess our method under diverse and difficult cross-domain scenarios.

\noindent\textbf{Data Augmentation}.
Data augmentation~\cite{wen2025rohoi, peng2024mitigating} is a key technique for the vision community. Traditional methods for object detection, like copy-paste~\cite{ghiasi2021simple}, cropping, and color jittering~\cite{bochkovskiy2020yolov4}, are simple but offer limited semantic variety. Recently, generative models—especially text-to-image models like ControlNet~\cite{zhang2023adding}, SDXL~\cite{podell2023sdxl}, FLUX~\cite{FLUX}, and FLUX-Fill~\cite{fluxfill2024} have enabled more advanced augmentations. Methods such as X-Paste~\cite{zhao2023x}, Lin et al.~\cite{lin2023explore}, and Zhang et al.~\cite{zhang2024advancing} generate new foregrounds to paste on diverse backgrounds, while others~\cite{ni2022imaginarynet, zhang2023diffusionengine, zhang2023adding, liu2023spatio, chen2023geodiffusion, zhou2023seeds, pan2025earthsynth, liu2025control} use text prompts to jointly create foregrounds and backgrounds. However, these methods typically rely on large amounts of in-domain data for training, which limits their adaptability to novel categories or unseen domains. In contrast, our Domain-RAG is training-free and leverages retrieved real-world images as visual priors to generate domain-consistent samples, making it well-suited for CD-FSOD.

\noindent\textbf{Retrieval-Augmented Generation in Vision}.
First introduced in NLP~\cite{lewis2020retrieval}, retrieval-augmented generation (RAG) enhances outputs by incorporating relevant retrieved content as external knowledge. Its strong performance has led to applications in vision tasks such as image captioning~\cite{ramos2023smallcap, li2024evcap}, visual question answering~\cite{lin2023fine, he2024g}, and image generation~\cite{blattmann2022retrieval, lyu2025realrag, zheng2025retrieval}, and pose estimation~\cite{wang2025rag}. However, current RAG-based image generation methods are aimed at open-ended synthesis and are not suited for object detection, particularly in cross-domain few-shot settings, where both domain alignment and object fidelity are crucial. To the best of our knowledge, we are the first to introduce a RAG-inspired, training-free image generation framework specifically designed for CD-FSOD.

\section{Proposed Method}
\label{method}

\noindent\textbf{Problem Setup}.
The CD-FSOD task aims to adapt an object detector from a source domain $\mathcal{D}_S$ to a target domain $\mathcal{D}_T$, where the data distributions $\mathcal{P}_S$ and $\mathcal{P}_T$ differ.
We use the few-shot setting, i.e., $N$-way $K$-shot protocol to evaluate detection results in $\mathcal{D}_T$.
Specifically, a support set $\mathcal{S^{ \textit{N} \times \textit{K}}} \subset \mathcal{D}_T$ provides $K$ labeled examples per novel class, and a query set $\mathcal{Q}$ is used for evaluation. 
We use the open-source setting introduced in the 1st CD-FSOD Challenge~\cite{fu2025ntire}, which allows foundation models pretrained on large-scale data, to explore the potential of foundation models in CD-FSOD.
Particularly, instead of pretraining on $\mathcal{D}_S$, we directly finetune a pretrained detector (e.g., GroundingDINO~\cite{liu2024grounding}) on the support set $\mathcal{S}$ and evaluate the results on the query set $\mathcal{Q}$. 
To mitigate the limited size of $\mathcal{S}$, we augment each support instance with $n$ synthetic images, effectively expanding each class from $K$ to $K \times (n+1)$ examples.

\begin{figure}[t]
    \vskip -0.15in
    \begin{center}
        \centerline{\includegraphics[width=1\columnwidth]{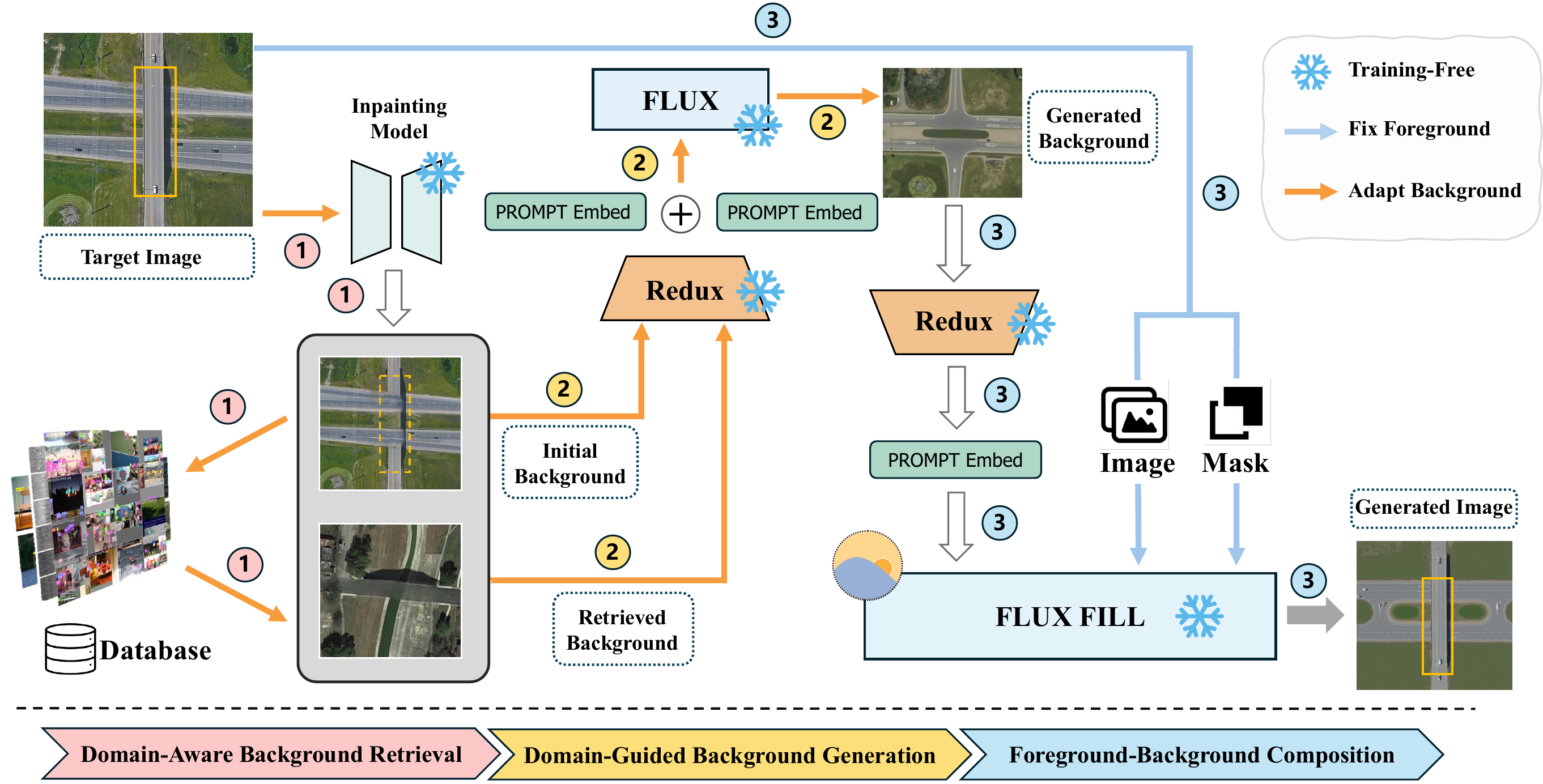}}
        \caption{\textbf{Illustration of our Domain-RAG.} Built on our principle of "fix the foreground, adapt the background", we first decompose image and process it with three key modules: domain-aware background retrieval, domain-guided background generation, and foreground-background composition.}
        \label{fig:framework}
    \end{center}
    \vskip -0.3in
\end{figure}

\noindent\textbf{Overview}. 
We propose Domain-RAG—a novel, training-free, retrieval-guided compositional generation framework that enhances support diversity by generating domain-aligned samples. To enable retrieval, we use COCO~\cite{lin2014microsoft} as the database $\mathcal{D}_{base}$, serving as a gallery of candidate backgrounds. Following the core principle of "\textit{fix the foreground, adapt the background}", Domain-RAG processes each support image $x \in \mathcal{S}$ by first decomposing it into foreground object(s) and background. As shown in Fig.~\ref{fig:framework}, the framework then proceeds through three key stages: \textit{1) domain-aware background retrieval} first obtains the inpainted background $b_{init}$ from $x$ and then retrieves $G$ candidate backgrounds $b_{re}$ from $\mathcal{D}_{base}$ that are semantically and stylistically similar. \textit{2) domain-guided background generation} feeds each $\{b_{init}, b_{re}\}$ pair into a generative model to synthesize a new domain-aligned background $b_{dom}$.
\textit{3) foreground-background composition} finally produces $n$ new images $x^+$  by compositing the preserved foreground onto each $b_{dom}$ using a mask-guided generative model.

\subsection{Domain-Aware Background Retrieval}
We propose a two-stage retrieval strategy that combines CLIP’s high-level semantic features with ResNet’s low-level style descriptors to search an existing image database.
The method retrieves images whose semantics and appearance are most similar to the target domain, providing background candidates that better match the target-domain distribution and thus enrich the support set $S$.

In practice, given a support image $x$, we remove the ground-truth bounding box with LaMa inpainting~\cite{suvorov2022resolution} to obtain a background without foreground $b_{init}$.
We use the CLIP encoder to extract embeddings from the initial background \( b_{init} \) and the database \( \mathcal{D}_{base} \), which we refer to as \( F_{bg} \) and \( F_{base} \), respectively.
We compute cosine similarity between the visual feature of the current background query $F_{bg}$ and the CLIP embedding of each sample in the database.
The top \( m \) most similar images are selected based on this similarity ranking, forming the candidate set \( B_{<clip, m>} \), which contains \( m \) images of the form \( b_{clip} \). The subscript notation indicates that the set is constructed using CLIP vision encoder and contains \( m \) elements.

Building on this step, we re-rank the $b_{clip}$ by extracting low-level style descriptors using shallow-layer ResNet features.
For each background image \( b_{clip} \) retrieved by CLIP, we extract its low-level feature map \( F \) using the early layers of a ResNet encoder.
We further compute the per-channel mean \( \mu_c \) and standard deviation \( \sigma_c \) by averaging over the spatial dimensions of the feature map $F$. 
Concatenating the means and standard deviations over all channels yields a 128-D style vector as,
\begin{equation}
    \mathbf{s}(b_{clip}) = \left[ \mu_1, \ldots, \mu_C, \, \sigma_1, \ldots, \sigma_C \right] \in \mathbb{R}^{2C}.
\end{equation}

For each retrieval candidate \( b_{clip} \) in the set \( B_{<clip, m>} \), we compute the style distance as the L2 norm between the style features of the original image \( b_{init} \) and the candidate:
\begin{equation}
    d = \left\| \mathbf{s}(b_{init}) - \mathbf{s}(b_{clip}) \right\|_2.
\end{equation}
Here, each \( d \) corresponds to a candidate in \( B_{<clip,\ m>} \), and we use the distance to rank and select the most stylistically similar backgrounds. We then re-rank the $m$ CLIP-retrieved candidates based on their style distances and retain the top $n$ images that are most similar in style. The resulting set of selected images is denoted as $B_{<re,\ n>}$.
The images indexed by $B_{<re,\ n>}$ serve as style-matched references for the subsequent background generation stage.

\subsection{Domain-Guided Background Generation}
To fully leverage the retrieved images while keeping the generation process training-free, we adopt the Flux-Redux model\cite{fluxredux2024} to encode each image into a prompt embedding.
Given our domain-aware retrieval results $B_{<re,\ n>}$, let $\text{redux}(\cdot)$ denote FLUX-Redux encoder and $\text{FLUX}(\cdot)$ denote the FLUX generator.
For each support image, we extract its clean background embedding $F_{bg} = \text{redux}(b_{init})$ and the embedding of the top retrieved image $b_{re}\in B_{<re,\ n>}$ as $F_{re}=\text{redux}(b_{re})$.
We then fuse them as $F_{dom} = \lambda_1 F_{bg} + \lambda_2 F_{re}$, where $\lambda_1$ and $ \lambda_2$ are hyper parameters.

Finally, the \text{FLUX} generator produces diverse background images at \(1024 \times 1024\) resolution by applying a generative function $\text{FLUX}$ to the domain embedding \( F_{dom} \) i.e.,  $b_{dom} ={\text{FLUX}}(F_{dom})$.
We sample this process \( n \) times to generate a set of diverse images \( \{ b^{(1)}, b^{(2)}, \dots, b^{(n)} \} \).

\subsection{Foreground-Background Composition}
Based on the diverse backgrounds generated in the previous stage, we aim to seamlessly integrate new backgrounds into the original images while preserving foreground pixels and maintaining the target-domain distribution.
To achieve this, we employ Flux-Fill for outpainting.
Specifically, for each corresponding support image \( x \), we construct a binary mask \( M \in \{0,1\}^{H \times W} \). The mask is computed based on the ground-truth bounding box \( \text{bbox}(x) \) as:
\begin{equation}
    M(p) = 
    \begin{cases}
        0, & \text{if } p \in \text{bbox}(x), \\
        1, & \text{otherwise}.
    \end{cases}
    \quad \text{for each } p \in \Omega_x,
\end{equation}
where \( \Omega_x \) denotes the spatial domain of image \( x \), and \( p \) indexes a pixel location.
We then extract the prompt embedding $F_{gen}$ by $F_{gen}=\texttt{redux}(b_{dom})$ and feed $\{x,M, F_{gen} \}$ into Flux-Fill.
To preserve foreground details, Flux-Fill encodes the input $x$ using a VAE and blends the encoded latent features with the initial noise. However, due to the VAE-based downsampling, it struggles to retain fine-grained structures such as small objects.
To mitigate this issue, before generation, we denote the up-sampling method $s_{up}$ on each image as,
\begin{equation}
    s_{up}(x) = 
    \begin{cases}
    0, & \text{if } \text{width}(x) > 1024 \text{ and } \text{height}(x) > 1024, \\
    1, & \text{otherwise}.
    \end{cases}
\end{equation}
After generation, we denote a corresponding down-sampling method $s_{down}$ as,
\begin{equation}
    s_{down}(x) = 
    \begin{cases}
    0, & \text{if } s_{up}(x) = 0, \\
    1, & \text{otherwise}.
    \end{cases}
\end{equation}
The model then repaints only the masked regions, merging the style of $b_{dom}$ while keeping the foreground object’s appearance and position unchanged. 
The final output of Domain-RAG, denoted $x^{+}$ , is given by,
\begin{equation}
    x^{+} = s_{down}\left(\text{Flux-Fill}\left( s_{up}(x), \, s_{up}(M), \, F_{gen} \right) \right).
\end{equation}
This completes the foreground-background composition, yielding an augmented support image with a domain-aligned background and unchanged foreground objects.

\subsection{Applying Domain-RAG to CD-FSOD}
In principle, our proposed \textbf{Domain-RAG} framework can be seamlessly integrated with any existing detector to enhance its performance in cross-domain scenarios.
As a training-free, plug-and-play data augmentation module, Domain-RAG requires no modification to the detection architecture or training pipeline.
Once the augmented support images are generated, the model is fine-tuned on the combination of the original support set $\mathcal{S}$ and the generated samples.
At inference time, Domain-RAG is not involved; the detector is evaluated directly on the original query set $\mathcal{Q}$.

\section{Experiments}
\label{exp}
\noindent\textbf{Setups}.
We conduct experiments on three FSOD tasks with domain shifts:
\textbf{1) CD-FSOD:} Following the CD-ViTO benchmark~\cite{fu2024cross}, we evaluate on six diverse target domains: ArTaxOr~\cite{GeirArTaxOr} (photorealistic), Clipart1k~\cite{inoue2018cross} (cartoon), DIOR~\cite{li2020object} (aerial), DeepFish~\cite{saleh2020realistic} (underwater), NEU-DET~\cite{song2013noise} (industrial), and UODD~\cite{jiang2021underwater} (underwater).
\textbf{2) Remote Sensing FSOD (RS-FSOD):} In addition to DIOR, we include NWPU VHR-10~\cite{niemeyer2014contextual}, a popular remote sensing dataset for FSOD.
\textbf{3) Camouflaged FSOD:} We also test on CAMO-FS~\cite{nguyen2024art}, a recent dataset with 47 categories where objects are deliberately camouflaged into the background.
For each task, we follow the standard dataset splits and evaluation protocols: 1/5/10-shot for CD-FSOD, 3/5/10/20-shot for RS-FSOD, and 1/2/3/5-shot for Camouflaged FSOD. Results are reported using mean Average Precision (mAP).

\begin{table}[h]
    \centering
    \vspace{-0.05in}
    \caption{
        \textbf{Main results (mAP) on the CD-FSOD benchmark} under the 1/5/10-shot setting. 
        $\dagger$ marks methods implemented or reproduced by us. Best results are highlighted in pink.
    }    
    \vspace{-0.05in}
    \label{tab:main}
    \resizebox{1.\columnwidth}{!}{
    \begin{tabular}{cllcccccccc}
    \toprule
    & \textbf{Method} & \textbf{Backbone} & \textbf{ArTaxOr} & \textbf{Clipart1k} &  \textbf{DIOR} &   \textbf{DeepFish}   &  \textbf{NEU-DET}  & \textbf{UODD}  &  \textbf{Average} \\
    \midrule
    \multirow{12}{*}{\rotatebox{90}{1-shot}} & Meta-RCNN  ~\cite{yan2019meta} & ResNet50 & 2.8 & - & 7.8 & - & - & 3.6 & / \\
    & TFA w/cos ~\cite{wang2020frustratingly} & ResNet50 & 3.1 & - & 8.0 & - & - & 4.4 & / \\
    & FSCE ~\cite{sun2021fsce} & ResNet50 &  3.7 & - & 8.6 & - & - & 3.9 & / \\ 
    & DeFRCN ~\cite{qiao2021defrcn} & ResNet50 & 3.6 & - & 9.3 & - & - & 4.5  & / \\ 
    & Distill-cdfsod ~\cite{xiong2023cd} & ResNet50 & 5.1 & 7.6 & 10.5 & nan &  nan &  5.9 & / \\
    \cline{2-10}
    & ViTDeT-FT ~\cite{li2022exploring} & ViT-B/14 & 5.9 & 6.1 & 12.9 & 0.9 & 2.4 & 4.0 & 5.4\\
    & Detic~\cite{zhou2022detecting} & ViT-L/14 & 0.6 & 11.4 & 0.1 & 0.9 & 0.0 & 0.0 & 2.2 \\
    & Detic-FT ~\cite{zhou2022detecting} & ViT-L/14 & 3.2 & 15.1 & 4.1 & 9.0 & 3.8 & 4.2 & 6.6 \\
    & DE-ViT~\cite{zhang2023detect} & ViT-L/14 & 0.4 & 0.5 & 2.7 & 0.4 & 0.4 & 1.5 & 1.0 \\
    & CD-ViTO~\cite{fu2024cross} & ViT-L/14 & 21.0 & 17.7 & 17.8 & 20.3 & 3.6 & 3.1 & 13.9 \\
    \cline{2-10} 
    & GroundingDINO$\dagger$~\cite{liu2024grounding} & Swin-B & 26.3 & 55.3 & 14.8 & 36.4 & 9.3 & 15.9 & 26.3 \\
    & ETS$\dagger$~\cite{pan2025enhance} & Swin-B & 28.1 & 55.8 & 12.7 & \cellcolor{pink!60}\textbf{39.3} & 11.7 & 18.9 & 27.8 \\
    & \textbf{Domain-RAG (Ours)} & Swin-B & \cellcolor{pink!60}\textbf{57.2} & \cellcolor{pink!60}\textbf{56.1} & \cellcolor{pink!60}\textbf{18.0} & 38.0 & \cellcolor{pink!60}\textbf{12.1} & \cellcolor{pink!60}\textbf{20.2} & \cellcolor{pink!60}\textbf{33.6} \\
    \midrule
    \multirow{12}{*}{\rotatebox{90}{5-shot}} 
    & Meta-RCNN  ~\cite{yan2019meta} & ResNet50 & 8.5 & - & 17.7 & - & - & 8.8 & / \\
    & TFA w/cos ~\cite{wang2020frustratingly} & ResNet50 & 8.8 & - & 18.1 & - & - & 8.7 & / \\
    & FSCE ~\cite{sun2021fsce} & ResNet50 & 10.2 & - & 18.7 & - & - & 9.6 & / \\ 
    & DeFRCN ~\cite{qiao2021defrcn} & ResNet50 &  9.9 & - & 18.9 & - & - & 9.9 & / \\ 
    & Distill-cdfsod ~\cite{xiong2023cd} & ResNet50 & 12.5 & 23.3 & 19.1 & 15.5 & 16.0 & 12.2 & 16.4 \\
    \cline{2-10}
    & ViTDeT-FT ~\cite{li2022exploring} & ViT-B/14 & 20.9 & 23.3 & 23.3 & 9.0 & 13.5 & 11.1 & 16.9 \\
    & Detic~\cite{zhou2022detecting} & ViT-L/14 & 0.6 & 11.4 & 0.1 & 0.9 & 0.0 & 0.0 & 2.2 \\
    & Detic-FT ~\cite{zhou2022detecting} & ViT-L/14 & 8.7 & 20.2 & 12.1 & 14.3 & 14.1 & 10.4 & 13.3 \\
    & DE-ViT~\cite{zhang2023detect} & ViT-L/14 & 10.1 & 5.5 & 7.8 & 2.5 & 1.5 & 3.1 & 5.1 \\
    & CD-ViTO~\cite{fu2024cross} & ViT-L/14 & 47.9 & 41.1 & 26.9 & 22.3 & 11.4 & 6.8 & 26.1 \\
    \cline{2-10} 
    & GroundingDINO$\dagger$~\cite{liu2024grounding} & Swin-B & 68.4 & 57.6 & 29.6 & 41.6 & 19.7 & 25.6 & 40.4 \\
    & ETS$\dagger$~\cite{pan2025enhance} & Swin-B & 64.5 & 59.7 & 29.3 & 42.1 & 23.5 & \cellcolor{pink!60}\textbf{27.7} & 41.1 \\
    & \textbf{Domain-RAG (Ours)} & Swin-B & \cellcolor{pink!60}\textbf{70.0} & \cellcolor{pink!60}\textbf{59.8} & \cellcolor{pink!60}\textbf{31.5} & \cellcolor{pink!60}\textbf{43.8} & \cellcolor{pink!60}\textbf{24.2} & 26.8 & \cellcolor{pink!60}\textbf{42.7} \\
    \midrule
    \multirow{12}{*}{\rotatebox{90}{10-shot}} 
    & Meta-RCNN  ~\cite{yan2019meta} & ResNet50 & 14.0 & - & 20.6 & - & - & 11.2 & / \\
    & TFA w/cos ~\cite{wang2020frustratingly} & ResNet50 & 14.8 & - & 20.5 & - & - & 11.8 & / \\
    & FSCE ~\cite{sun2021fsce} & ResNet50 & 15.9 & - & 21.9 & - & - & 12.0 & / \\ 
    & DeFRCN ~\cite{qiao2021defrcn} & ResNet50 & 15.5 & - & 22.9 & - & - & 12.1 & / \\ 
    & Distill-cdfsod ~\cite{xiong2023cd} & ResNet50 & 18.1 & 27.3 & 26.5 & 15.5 & 21.1 & 14.5 & 20.5 \\ 
    \cline{2-10}
    & ViTDeT-FT ~\cite{li2022exploring} & ViT-B/14 & 23.4 & 25.6 & 29.4 & 6.5 & 15.8 & 15.6 & 19.4 \\ 
    & Detic~\cite{zhou2022detecting} & ViT-L/14 & 0.6 & 11.4 & 0.1 & 0.9 & 0.0 & 0.0 & 2.2 \\
    & Detic-FT ~\cite{zhou2022detecting} & ViT-L/14 & 12.0 & 22.3 & 15.4 & 17.9 & 16.8 & 14.4 & 16.5 \\
    & DE-ViT~\cite{zhang2023detect} & ViT-L/14 & 9.2 & 11.0 & 8.4 & 2.1 & 1.8 & 3.1 & 5.9 \\
    & CD-ViTO~\cite{fu2024cross} & ViT-L/14 & 60.5 & 44.3 & 30.8 & 22.3 & 12.8 & 7.0 & 29.6 \\
    \cline{2-10} 
    & GroundingDINO$\dagger$~\cite{liu2024grounding} & Swin-B & 73.0 & 58.6 & 37.2 & 38.5 & 25.5 & 30.3 & 43.9 \\
    & ETS$\dagger$~\cite{pan2025enhance} & Swin-B & 70.6 & 60.8 & 37.5 & \cellcolor{pink!60}\textbf{42.8} & 26.1 & 28.3 & 44.4 \\
    & \textbf{Domain-RAG (Ours)} & Swin-B & \cellcolor{pink!60}\textbf{73.4} & \cellcolor{pink!60}\textbf{61.1} & \cellcolor{pink!60}\textbf{39.0} & 41.3 & \cellcolor{pink!60}\textbf{26.3} & \cellcolor{pink!60}\textbf{31.2} & \cellcolor{pink!60}\textbf{45.4} \\
    \bottomrule
    \end{tabular}}
    \vspace{-0.2in}
\end{table}

\vspace{1.5em}132

\textbf{Implementation Details.}
We use pretrained GroundingDINO~\cite{liu2024grounding} with Swin-Transformer~\cite{liu2021swin} Base (Swin-B) as backbone as our baseline.
For the retrieval stage, the hyper parameters are set to \( m =100\), $n=5$, throughout all experiments. For the background generation stage, fusion hyper parameters $\lambda_1$ and $ \lambda_2$ are  set to 1.0 and 0.8 respectively.
We fine-tune the model for $30$ epochs by default, but reduce to $5$ for faster-converging datasets like Clipart1k and DeepFish. We use AdamW~\cite{loshchilov2017decoupled} with learning rate and weight decay set to $1 \times 10^{-4}$, and we scale the backbone's learning rate by $0.1$. All experiments are run on four Tesla V100 GPUs or eight 5880 Ada GPUs, or a single A800 GPU. Further details are in the Appendix.

\subsection{Main Comparison Results}

\noindent\textbf{CD-FSOD Results}. 
Tab.~\ref{tab:main} summarizes the main comparison results on CD-FSOD under 1/5/10 shots across six novel targets. 
Particularly, we include several competitors: Meta-RCNN~\cite{yan2019meta}, TFA w/cos~\cite{wang2020frustratingly}, FSCE~\cite{sun2021fsce}, DeFRCN~\cite{qiao2021defrcn}, Distill-cdfsod~\cite{xiong2023cd}, ViTDeT-FT~\cite{li2022exploring}, Detic/Detic-FT~\cite{zhou2022detecting}, DE-ViT~\cite{zhang2023detect} as reported in CD-ViTO~\cite{fu2024cross}.
In addition, we also report the results of fine-tuned GroundingDINO~\cite{liu2024grounding}, ETS~\cite{pan2025enhance} to compare with our Domain-RAG.
Note that both ETS and Domain-RAG build on GroundingDINO but with different augmentation strategies.

We highlight that Domain-RAG consistently outperforms existing competitors across most target domains, achieving new state-of-the-art (SOTA) results.
Compared to the GroundingDINO baseline, our method improves mAP by $7.3$, $2.3$, and $1.5$ points under the 1, 5, and 10 shots, respectively.
These results not only show the effectiveness of Domain-RAG, but also reveal its superiority over other proposed augmentation strategies such as ETS.
Beyond the average gains, we also notice:
1) \textit{Significant gains on ArTaxOr.}
Domain-RAG achieves a $117.5\%$ relative improvement in the 1-shot setting.
We attribute this to the strong semantic and visual compatibility between ArTaxOr and the retrieved COCO-style backgrounds, where ArTaxOr features the fine-grained foreground but with a relatively close visual domain to COCO regarding background.
2) \textit{Robustness under low-shot settings.}
The largest gains are observed in the 1-shot scenario, which is the most challenging FSOD scenario.
This shows our benefits under severe data scarcity.
3) \textit{Strong generalization to severe domain shift.}
On NEU-DET, an industrial defect detection dataset characterized by uncommon objects and background styles, Domain-RAG consistently improves all shot settings, demonstrating its capability to handle the most challenging cross-domain FSOD cases.

\noindent\textbf{RS-FSOD Results.}
Tab.~\ref{tab:nwpu_results} summarizes the results on the NWPU VHR-10 remote sensing dataset~\cite{niemeyer2014contextual} under the 3/5/10/20-shot settings.
The dataset is divided into $7$ base classes and $3$ novel classes.
The table is split into two parts.
In the \textit{upper part}, we follow the standard RS-FSOD protocol: models are first trained on the base classes and then fine-tuned and evaluated on the novel classes.
Under this setting, the base classes contain a sufficient number of annotated samples, and we apply our augmentation strategy on top of the previous state-of-the-art method SEA-FSDet~\cite{liu2024few}, and report the mean Average Precision (mAP) over the 3 novel classes.
In the \textit{lower part}, we explore the dataset in a CD-FSOD setting, where the pretrained model is directly fine-tuned on all 10 classes (both base and novel), each with only a few labeled samples.
To ensure comparability with the upper setting, the reported mAP here reflects performance exclusively on the three novel categories.

\vspace{0.5em}

\begin{table}[h]
\vspace{-0.1in}
\centering
\caption{\textbf{Main results (mAP) on the NWPU VHR-10  benchmark} under the 3/5/10/20-shot settings. The upper part follows the standard \textbf{RS-FSOD} problem setup, while the lower part adapts 
 \textbf{CD-FSOD} setup, with the best results highlighted in pink. $\dagger$ means results are produced by us.}
\label{tab:nwpu_results}
\scriptsize 
\begin{tabular}{lccccccc}
\toprule
\textbf{Method} & \textbf{Training Setting} & \textbf{Backbone} & \textbf{3-shot} & \textbf{5-shot} & \textbf{10-shot} & \textbf{20-shot} & \textbf{Average} \\
\midrule
Meta-RCNN~\cite{yan2019meta} & RS-FSOD & ResNet-50 & 20.51 & 21.77 & 26.98 & 28.24 & 24.38 \\
FsDetView~\cite{kang2019few} & RS-FSOD & ResNet-50 & 24.56 & 29.55 & 31.77 & 32.73 & 29.65 \\
TFA w/cos~\cite{wang2020frustratingly} & RS-FSOD & ResNet-50 & 16.17 & 20.49 & 21.22 & 21.57 & 19.86 \\
\textbf{P-CNN}~\cite{cheng2021prototype} & RS-FSOD & ResNet-50 & 41.80 & 49.17 & 63.29 & 66.83 & 55.27 \\
FSOD~\cite{fan2020few} & RS-FSOD & ResNet-50 & 10.95 & 15.13 & 16.23 & 17.11 & 14.86 \\
FSCE~\cite{sun2021fsce} & RS-FSOD & ResNet-50 & 41.63 & 48.80 & 59.97 & 79.60 & 57.50 \\
ICPE~\cite{lu2023breaking} & RS-FSOD & ResNet-50 & 6.10 & 9.10 & 12.00 & 12.20 & 9.85 \\
VFA~\cite{han2023few} & RS-FSOD & ResNet-50 & 13.14 & 15.08 & 13.89 & 20.18 & 15.57 \\
SAE-FSDet~\cite{liu2024few} & RS-FSOD & ResNet-50 & 57.96 & 59.40 & 71.02 & \cellcolor{pink!60}\textbf{85.08} & 68.36 \\
\textbf{Domain-RAG (Ours)} & RS-FSOD & ResNet-50 & \cellcolor{pink!60}\textbf{59.99} & \cellcolor{pink!60}\textbf{65.78} & \cellcolor{pink!60}\textbf{72.87} & 84.05 & \cellcolor{pink!60}\textbf{70.67} \\
\midrule
GroundingDINO$\dagger$~\cite{liu2024grounding} & CD-FSOD & Swin-B & 57.1 & 61.3 & 65.1 & 69.5 & 63.3 \\
\textbf{Domain-RAG (Ours)} & CD-FSOD & Swin-B & \cellcolor{pink!60}\textbf{58.2} & \cellcolor{pink!60}\textbf{62.1} & \cellcolor{pink!60}\textbf{66.6} & \cellcolor{pink!60}\textbf{69.7} & \cellcolor{pink!60}\textbf{64.2} \\
\bottomrule
\end{tabular}
\vspace{-0.1in}
\end{table}

\vspace{0.5em}

Notably, from the upper standard RS-FSOD results, we observe the following findings: 1) Our Domain-RAG achieves the best result via improving the strong SEA-FSDet, achieving $2.31$ mAP improvement across all shots on average.
This indicates that our plug-and-play augmented method is compatible with existing methods.
2) Minor decrease is observed for 20-shot, from $85.08$ to $84.05$.
We speculate that it is due to the base training being sufficient. Further augmentation in this regime may lead to overfitting on synthetic data patterns rather than benefiting novel-class generalization. 
From the lower part of the CD-FSOD setting results, we highlight that our method again improves the strong GroundingDINO baseline, indicating its effectiveness.

\noindent\textbf{Camouflaged FSOD Results.}
Tab.~\ref{tab:comparison_shot_settings} presents the results on the CAMO-FS~\cite{nguyen2024art} under 1/2/3/5 shots.
All categories in this dataset are treated as novel classes and are further split into a support set and a query set, naturally aligning with the formulation of CD-FSOD.
The first two rows in the table report the results of "FS-CDIS-ITL" and "FS-CDIS-IMS", two methods developed from the original CAMO-FS paper.
Below that, we include our reproduced baseline using GroundingDINO as the detector, along with the results of our proposed Domain-RAG method built on top of GroundingDINO.

As shown by the results, the large-scale pretrained model, i.e., GroundingDINO, brings a substantial performance boost to this task, improving results from around $7$ to over $65$ mAP.

We believe this remarkable advancement will advance the frontier of this field.
Moreover, the performance gains introduced by our proposed method over the GroundingDINO baseline remain consistently clear across all shot settings.
The consistent success across CD-FSOD, RS-FSOD, and camouflaged FSOD, covering eight challenging and diverse domains, demonstrates that our method serves as a general and effective solution for addressing the gap issue in few-shot object detection.

\begin{table}[t]
\centering
\caption{
\textbf{Main results (mAP) on the  Camouflage FSOD} under the 1/2/3/5-shot settings. $\dagger$ means the results are produced by us, the best results are highlighted in pink.}
\label{tab:comparison_shot_settings}
\scriptsize 
\begin{tabular}{lcccccc}
\toprule
\textbf{Method} & \textbf{Backbone} & \textbf{1-shot} & \textbf{2-shot} & \textbf{3-shot} & \textbf{5-shot} & \textbf{Average} \\
\midrule
FS-CDIS-ITL~\cite{nguyen2024art} & ResNet-101  & 4.0 & 7.3 & 7.5 & 9.8 & 7.1 \\
FS-CDIS-IMS~\cite{nguyen2024art} & ResNet-101  & 4.5 & 7.0 & 7.6 & 10.4 & 7.4 \\
GroundingDINO$\dagger$~\cite{liu2024grounding}  & Swin-B & 63.4 & 66.8 & 67.1 & 69.1 & 66.6 \\
\textbf{Domain-RAG (Ours)} & Swin-B & \cellcolor{pink!60}\textbf{65.5} & \cellcolor{pink!60}\textbf{67.7} & \cellcolor{pink!60}\textbf{68.3} & \cellcolor{pink!60}\textbf{70.3} & \cellcolor{pink!60}\textbf{68.0} \\
\bottomrule
\end{tabular}
\vspace{-0.2in}
\end{table} 

\vspace{0.5em}

\subsection{Comparison with Other Augmentation Methods}
To assess the effectiveness of our Domain-RAG framework, we compare it with several strong baselines that are designed for augmenting data for CD-FSOD.
Specifically, 1) "Copy-Paste" directly overlays foreground objects onto random COCO backgrounds without considering semantic relevance or compositional integrity.
2) "Foreground Augmentation" attempts to diversify object appearances by inpainting new foregrounds after object removal.
This is done by using the category label of each bounding box as a text prompt and applying SDXL-inpaint to generate a new foreground after removing the original object.
3) "Background Augmentation", which we use InstructBLIP~\cite{instructblip} to caption the remaining background, and guide SDXL to generate a new background based on this caption after removing the foreground from a target image.
To ensure fair comparison, all the augmentation methods use $G=5$.
Comparison results are summarized in Tab.~\ref{tab:aug}.
The results are reported on CD-FSOD under the 1-shot setting.

\vspace{0.5em}

\begin{table}[h]
\centering
\normalsize
\vspace{-0.1in}
\caption{Comparison of augmentation methods (mAP) on the CD-FSOD benchmark under 1-shot.}
\vspace{-0.05in}
\label{tab:aug}
\resizebox{1\textwidth}{!}{
\begin{tabular}{lccccccc}
\toprule
\textbf{Method} & \textbf{ArTaxOr} & \textbf{Clipart} & \textbf{DIOR} & \textbf{DeepFish} & \textbf{NEU-DET} & \textbf{UODD} & \textbf{Average}\\
\midrule
GroundingDINO & 26.3 & 55.3 & 14.8 & 36.4 & 9.3 & 15.9 & 26.3 \\
Copy-Paste & 38.8 & 55.0 & 15.0 & 36.4 & 8.4 & 14.2 & 27.9 \\
Foreground Augmentation  & 32.4 & 56.1 & 13.9 & \cellcolor{pink!60}\textbf{41.4} & 9.6 & 14.9 & 28.1 \\
Background Augmentation & 52.3 & 53.7 & 16.9 & 34.2 & 8.9 & 10.8 & 29.5 \\
\textbf{Domain-RAG (Ours)} & \cellcolor{pink!60}\textbf{57.2} & \cellcolor{pink!60}\textbf{56.1} & \cellcolor{pink!60}\textbf{18.0} & 38.0 & \cellcolor{pink!60}\textbf{12.1} & \cellcolor{pink!60}\textbf{20.2} & \cellcolor{pink!60}\textbf{33.6} \\
\bottomrule
\end{tabular}}
\vspace{-0.05in}
\end{table}

\vspace{3em}

We observe that: 1) copy-paste methods can work reasonably well on relatively simple datasets such as ArTaxOr.
However, due to a lack of semantic consistency and domain alignment, they tend to fail on most target domains.
2) Foreground-augmentation baseline performs well when the foreground is visually simple and isolated, for example, in datasets like DeepFish, where only a single object is present.
However, due to the potential semantic shift issue, it failed on more complex datasets such as DIOR and UODD.
3) Background-augmentation baseline also suffers in CD-FSOD, often failing on datasets with distinctive domain characteristics, such as NEU-DET.
4) In contrast, our method consistently improves upon the baseline across all datasets, demonstrating its robustness and effectively addressing the limitations of prior approaches.

\subsection{More Analysis}
\label{sec:more-analysis}
\noindent\textbf{Ablation Study on Proposed Modules.}
To evaluate each module's effectiveness, we conduct ablation studies by removing or replacing it with naive alternatives. As a typical challenging case, the NEU-DET under a 1-shot setting is demonstrated as an example.
Results are shown in Fig.~\ref{fig:abla} (a).
Specifically, 1) the grey bar marks the "baseline", i.e., vanilla fine-tuned GroundingDINO.
2) The pink bar ("w/o background retrieval") disables the domain-aware retrieval module and replaces the backgrounds with random COCO images while keeping the rest of the pipeline unchanged.
3) The yellow bar ("w/o background generation") skips the domain-guided background generation and directly performs the foreground-background composition with the raw retrieved images from COCO.
4) The blue bar ("copy-paste as compositional") removes the last foreground–background composition stage and simply pastes the foreground onto the domain-aligned generated background.
5) The last colorful bar represents our full Domain-RAG.

Results show that our full model outperforms all ablated variants, achieving the best overall performance.
Furthermore, we observe the following: 1) By comparing our method with the pink bar, we verify that the domain-aware background retrieval stage provides backgrounds that are better aligned with the target domain.
2) The comparison between the gray and yellow bars indicates that simply augmenting backgrounds using COCO images offers limited benefits.
In contrast, the domain-guided background generation stage significantly improves performance by producing backgrounds that are both semantically and stylistically aligned, as evidenced by the gap between the yellow and final colorful bars.
3) The performance drop seen with the blue bar underscores the importance of the foreground-background composition stage, which enables seamless integration of foreground objects into the generated backgrounds.
Together, these observations confirm that each component of Domain-RAG is both indispensable and complementary for achieving robust CD-FSOD performance.

\noindent\textbf{The Construction of RAG Database}
In the defined (closed-source) CD-FSOD setting as proposed in CD-ViTO~\cite{fu2024cross}, COCO serves as the only single-source dataset for training, while other datasets (ArTaxOr, Clipart1k, DIOR, DeepFish, NEU-DET, UODD) are treated as unseen targets. 
Using COCO as the RAG database brings two key advantages: 
(1) It does not introduce any extra data beyond the default setting, ensuring the fairness of comparison; 
(2) COCO provides diverse and general-domain backgrounds that better cover novel domain scenarios.

Furthermore, we conducted additional experiments using different database options, including COCO with reduced category numbers, NEU-DET (non-general-domain), and miniImageNet. 

\vspace{0.5em}
\begin{table}[htbp]
\centering
\setlength{\tabcolsep}{3.5pt}
\renewcommand{\arraystretch}{0.9}
\caption{Effect of different database choices on Domain-RAG performance.}
\label{tab:db_quantity}
\resizebox{\linewidth}{!}{
\begin{tabular}{lccccccc}
\toprule
\textbf{DataBase} & \textbf{ArTaxOr} & \textbf{Clipart1k} & \textbf{DIOR} & \textbf{FISH} & \textbf{NEU-DET} & \textbf{UODD} & \textbf{Avg} \\
\midrule
Base (GroundingDINO) & 26.3 & 55.3 & 14.8 & 36.4 & 9.3 & 15.9 & 26.3 \\
COCO-1class & 50.1 & 55.0 & 15.7 & 36.6 & 12.0 & 16.0 & 30.9 \\
COCO-5classes & 51.0 & 55.1 & 16.6 & 36.7 & 11.9 & 17.5 & 31.5 \\
COCO-20classes & 53.0 & 56.2 & 16.2 & 37.0 & 12.2 & 18.9 & 32.3 \\
\textbf{COCO-80classes (Ours)} &  \cellcolor{pink!60}\textbf{57.2} &  \cellcolor{pink!60}\textbf{56.1} &  \cellcolor{pink!60}\textbf{18.0} &  \cellcolor{pink!60}\textbf{38.0} &  12.1 &  \cellcolor{pink!60}\textbf{20.2} &  \cellcolor{pink!60}\textbf{33.6} \\
NEU-DET & 49.8 & 55.2 & 16.4 & 37.0 & 12.0 & 16.1 & 31.1 \\
miniImageNet & 55.6 & 53.2 & 15.6 & 38.0 & \cellcolor{pink!60}\textbf{14.0} & 16.2 & 32.1 \\
\bottomrule
\end{tabular}}
\end{table}

\vspace{1em}

From the results summarized in Table~\ref{tab:db_quantity}, we observe that: 
(1) broader category coverage consistently improves performance; 
(2) general-domain databases such as COCO outperform specific-domain ones like NEU-DET; 
and (3) although miniImageNet can serve as an alternative database, it performs slightly worse than COCO due to its larger foreground regions and less diverse backgrounds. 
These findings demonstrate that our Domain-RAG consistently enhances the base GroundingDINO across all benchmarks, validating the robustness and effectiveness of our approach.

\noindent\textbf{Visualization of Generation Images.}  
To provide a more intuitive illustration of our method’s effectiveness, we present qualitative results in Fig.~\ref{fig:abla} (b).
Each example shows the original target image from a different domain in the first row and the corresponding generated image in the second row, with annotated object bounding boxes.
From the results, we observe the following: 1) Our method effectively preserves the foreground object without introducing noticeable changes, even in challenging domains such as remote sensing, underwater, and industrial defect scenarios.
2) The generated images successfully introduce new backgrounds while maintaining overall semantic coherence and visual consistency with the original domain.
Also, the outputs appear natural and realistic. These two observations align well with our goals and further validate the effectiveness of the proposed Domain-RAG framework. 

\begin{figure}[t]
    \vskip -0.1in
    \centering
    \centerline{\includegraphics[width=\columnwidth]{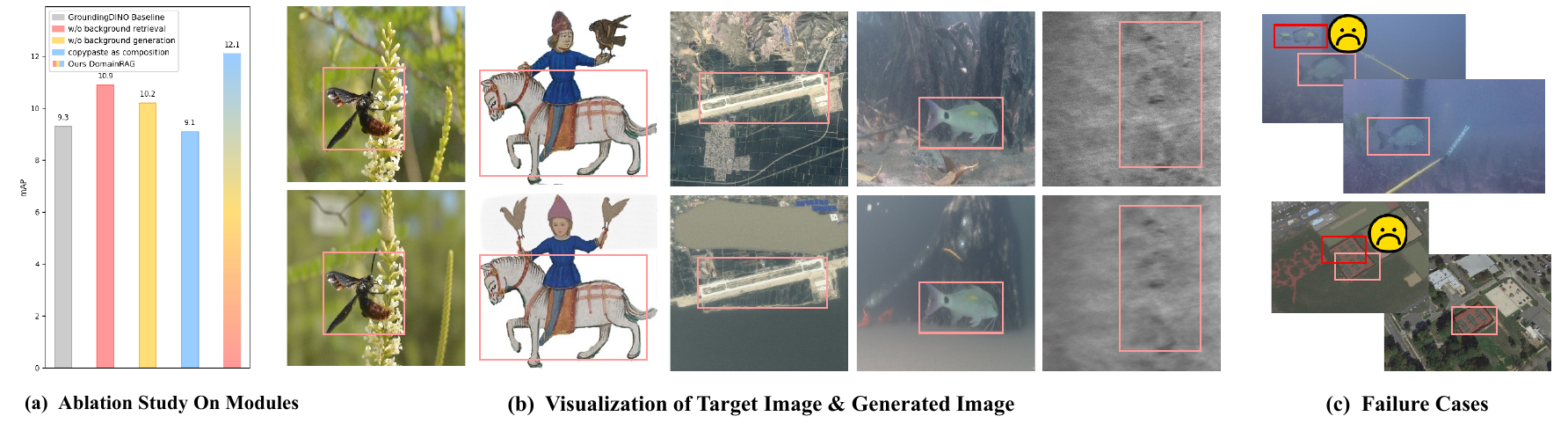}}
        \vskip -0.1in
    \caption{(a) Ablation study on modules, results reported on NEU-DET, 1 shot. (b) Visualization of target image (top row) and generated image (second row). (c) Failure Cases.}
    \label{fig:abla}
    \vskip -0.1in
\end{figure}

\noindent\textbf{Failure Cases and Limitations.} 
We further examine the quality of the generated images and observe that, in a few cases, our model exhibits foreground information leakage.
Parts of the foreground object are unintentionally regenerated within the background, as illustrated in the area highlighted with red boxes of Fig.~\ref{fig:abla} (c).
Since these regenerated foregrounds are not explicitly controlled and lack corresponding annotations, they may introduce noise into model fine-tuning and potentially harm detection performance.

Additional results, including ablation studies of our proposed modules on other targets, more detailed analyses, and extended visualizations, are provided in the Appendix.

\section{Conclusion}
In this paper, we investigate few-shot object detection (FSOD) across domains—a more realistic yet significantly more challenging scenario than conventional FSOD.
We focus on three representative tasks: cross-domain FSOD (CD-FSOD), remote sensing FSOD (RS-FSOD), and camouflaged FSOD.
To improve performance under these settings, we propose \textbf{Domain-RAG}, a training-free compositional image generation framework designed to produce domain-aligned and detection-friendly samples.
Unlike existing text-to-image generation approaches that rely solely on textual prompts, Domain-RAG retrieves semantically and stylistically similar images as structured priors to guide the generation process.
To the best of our knowledge, this is the first application of retrieval-augmented generation to cross-domain object detection, particularly in a training-free way suitable for low-shot scenarios.
Domain-RAG achieves new state-of-the-art results across all three tasks, demonstrating its generalization ability and opening new directions for training-free data synthesis.

\section{Acknowledgment}

This work was supported by the Science and Technology Commission of Shanghai Municipality (No. 24511103100).The authors gratefully thank the organization for their support and resources.


\clearpage
\bibliographystyle{plain}
\bibliography{reference}

@inproceedings{zhang2024advancing,
  title={Advancing Controllable Diffusion Model for Few-Shot Object Detection in Optical Remote Sensing Imagery},
  author={Zhang, Tong and Zhuang, Yin and Zhang, Xinyi and Wang, Guanqun and Chen, He and Bi, Fukun},
  booktitle={IGARSS 2024-2024 IEEE International Geoscience and Remote Sensing Symposium},
  pages={7600--7603},
  year={2024},
  organization={IEEE}
}

@inproceedings{zhao2023x,
  title={X-paste: Revisiting scalable copy-paste for instance segmentation using clip and stablediffusion},
  author={Zhao, Hanqing and Sheng, Dianmo and Bao, Jianmin and Chen, Dongdong and Chen, Dong and Wen, Fang and Yuan, Lu and Liu, Ce and Zhou, Wenbo and Chu, Qi and others},
  booktitle={International Conference on Machine Learning},
  pages={42098--42109},
  year={2023},
  organization={PMLR}
}

@inproceedings{lin2023explore,
  title={Explore the power of synthetic data on few-shot object detection},
  author={Lin, Shaobo and Wang, Kun and Zeng, Xingyu and Zhao, Rui},
  booktitle={Proceedings of the IEEE/CVF conference on computer vision and pattern recognition},
  pages={638--647},
  year={2023}
}

@article{ni2022imaginarynet,
  title={Imaginarynet: Learning object detectors without real images and annotations},
  author={Ni, Minheng and Huang, Zitong and Feng, Kailai and Zuo, Wangmeng},
  journal={arXiv preprint arXiv:2210.06886},
  year={2022}
}

@article{chen2023geodiffusion,
  title={Geodiffusion: Text-prompted geometric control for object detection data generation},
  author={Chen, Kai and Xie, Enze and Chen, Zhe and Wang, Yibo and Hong, Lanqing and Li, Zhenguo and Yeung, Dit-Yan},
  journal={arXiv preprint arXiv:2306.04607},
  year={2023}
}

@article{qiao2021defrcn,
  title={DeFRCN: Decoupled Faster R-CNN for Few-Shot Object Detection},
  author={Qiao, Limeng and Zhao, Yuxuan and Li, Zhiyuan and Qiu, Xi and Wu, Jianan and Zhang, Chi},
  journal={arXiv preprint arXiv:2108.09017},
  year={2021}
}

@article{zhang2023detect,
  title={Detect everything with few examples},
  author={Zhang, Xinyu and Liu, Yuhan and Wang, Yuting and Boularias, Abdeslam},
  journal={arXiv preprint arXiv:2309.12969},
  year={2023}
}

@article{wang2020frustratingly,
  title={Frustratingly simple few-shot object detection},
  author={Wang, Xin and Huang, Thomas E and Darrell, Trevor and Gonzalez, Joseph E and Yu, Fisher},
  journal={arXiv preprint arXiv:2003.06957},
  year={2020}
}

@inproceedings{sun2021fsce,
  title={Fsce: Few-shot object detection via contrastive proposal encoding},
  author={Sun, Bo and Li, Banghuai and Cai, Shengcai and Yuan, Ye and Zhang, Chi},
  booktitle={computer vision and pattern recognition},
  year={2021}
}

@inproceedings{kang2019few,
  title={Few-shot object detection via feature reweighting},
  author={Kang, Bingyi and Liu, Zhuang and Wang, Xin and Yu, Fisher and Feng, Jiashi and Darrell, Trevor},
  booktitle={Proceedings of the IEEE/CVF international conference on computer vision},
  pages={8420--8429},
  year={2019}
}

@misc{instructblip,
      title={InstructBLIP: Towards General-purpose Vision-Language Models with Instruction Tuning}, 
      author={Wenliang Dai and Junnan Li and Dongxu Li and Anthony Meng Huat Tiong and Junqi Zhao and Weisheng Wang and Boyang Li and Pascale Fung and Steven Hoi},
      year={2023},
      eprint={2305.06500},
      archivePrefix={arXiv},
      primaryClass={cs.CV}
}

@inproceedings{xiong2023cd,
  title={CD-FSOD: A benchmark for cross-domain few-shot object detection},
  author={Xiong, Wuti},
  booktitle={ICASSP 2023-2023 IEEE International Conference on Acoustics, Speech and Signal Processing (ICASSP)},
  pages={1--5},
  year={2023},
  organization={IEEE}
}

@inproceedings{GeirArTaxOr,
  title={Arthropod taxonomy orders object detection dataset},
  author={Geir Drange},
  booktitle={https://doi.org/10.34740/kaggle/dsv/1240192},
  year={2019}
}

@inproceedings{inoue2018cross,
  title={Cross-domain weakly-supervised object detection through progressive domain adaptation},
  author={Inoue, Naoto and Furuta, Ryosuke and Yamasaki, Toshihiko and Aizawa, Kiyoharu},
  booktitle={CVPR},
  year={2018}
}

@article{li2020object,
  title={Object detection in optical remote sensing images: A survey and a new benchmark},
  author={Li, Ke and Wan, Gang and Cheng, Gong and Meng, Liqiu and Han, Junwei},
  journal={ISPRS},
  year={2020}
}

@article{saleh2020realistic,
  title={A realistic fish-habitat dataset to evaluate algorithms for underwater visual analysis},
  author={Saleh, Alzayat and Laradji, Issam H and Konovalov, Dmitry A and Bradley, Michael and Vazquez, David and Sheaves, Marcus},
  journal={Scientific Reports},
  year={2020}
}

@article{song2013noise,
  title={A noise robust method based on completed local binary patterns for hot-rolled steel strip surface defects},
  author={Song, Kechen and Yan, Yunhui},
  journal={Applied Surface Science},
  year={2013}
}

@inproceedings{jiang2021underwater,
  title={Underwater species detection using channel sharpening attention},
  author={Jiang, Lihao and Wang, Yi and Jia, Qi and Xu, Shengwei and Liu, Yu and Fan, Xin and Li, Haojie and Liu, Risheng and Xue, Xinwei and Wang, Ruili},
  booktitle={ACM MM},
  year={2021}
}

@article{nguyen2024art,
  title={The Art of Camouflage: Few-Shot Learning for Animal Detection and Segmentation},
  author={Nguyen, Thanh-Danh and Vu, Anh-Khoa Nguyen and Nguyen, Nhat-Duy and Nguyen, Vinh-Tiep and Ngo, Thanh Duc and Do, Thanh-Toan and Tran, Minh-Triet and Nguyen, Tam V},
  journal={IEEE Access},
  year={2024},
  publisher={IEEE}
}

@article{niemeyer2014contextual,
  title={Contextual classification of lidar data and building object detection in urban areas},
  author={Niemeyer, Joachim and Rottensteiner, Franz and Soergel, Uwe},
  journal={ISPRS journal of photogrammetry and remote sensing},
  volume={87},
  pages={152--165},
  year={2014},
  publisher={Elsevier}
}

@inproceedings{yan2019meta,
  title={Meta r-cnn: Towards general solver for instance-level low-shot learning},
  author={Yan, Xiaopeng and Chen, Ziliang and Xu, Anni and Wang, Xiaoxi and Liang, Xiaodan and Lin, Liang},
  booktitle={Proceedings of the IEEE/CVF international conference on computer vision},
  pages={9577--9586},
  year={2019}
}

@inproceedings{li2022exploring,
  title={Exploring plain vision transformer backbones for object detection},
  author={Li, Yanghao and Mao, Hanzi and Girshick, Ross and He, Kaiming},
  booktitle={ECCV},
  year={2022}
}

@inproceedings{zhou2022detecting,
  title={Detecting twenty-thousand classes using image-level supervision},
  author={Zhou, Xingyi and Girdhar, Rohit and Joulin, Armand and Kr{\"a}henb{\"u}hl, Philipp and Misra, Ishan},
  booktitle={ECCV},
  year={2022}
}

@inproceedings{liu2024grounding,
  title={Grounding dino: Marrying dino with grounded pre-training for open-set object detection},
  author={Liu, Shilong and Zeng, Zhaoyang and Ren, Tianhe and Li, Feng and Zhang, Hao and Yang, Jie and Jiang, Qing and Li, Chunyuan and Yang, Jianwei and Su, Hang and others},
  booktitle={European Conference on Computer Vision},
  pages={38--55},
  year={2024},
  organization={Springer}
}

@article{liu2024few,
  title={Few-shot object detection in remote sensing images via label-consistent classifier and gradual regression},
  author={Liu, Yanxing and Pan, Zongxu and Yang, Jianwei and Zhang, Bingchen and Zhou, Guangyao and Hu, Yuxin and Ye, Qixiang},
  journal={IEEE Transactions on Geoscience and Remote Sensing},
  year={2024},
  publisher={IEEE}
}

@article{lyu2025realrag,
  title={RealRAG: Retrieval-augmented Realistic Image Generation via Self-reflective Contrastive Learning},
  author={Lyu, Yuanhuiyi and Zheng, Xu and Jiang, Lutao and Yan, Yibo and Zou, Xin and Zhou, Huiyu and Zhang, Linfeng and Hu, Xuming},
  journal={arXiv preprint arXiv:2502.00848},
  year={2025}
}

@article{blattmann2022retrieval,
  title={Retrieval-augmented diffusion models},
  author={Blattmann, Andreas and Rombach, Robin and Oktay, Kaan and M{\"u}ller, Jonas and Ommer, Bj{\"o}rn},
  journal={Advances in Neural Information Processing Systems},
  volume={35},
  pages={15309--15324},
  year={2022}
}

@article{kohler2023few,
  title={Few-shot object detection: A comprehensive survey},
  author={K{\"o}hler, Mona and Eisenbach, Markus and Gross, Horst-Michael},
  journal={IEEE Transactions on Neural Networks and Learning Systems},
  year={2023}
}

@inproceedings{fu2024cross,
  title={Cross-domain few-shot object detection via enhanced open-set object detector},
  author={Fu, Yuqian and Wang, Yu and Pan, Yixuan and Huai, Lian and Qiu, Xingyu and Shangguan, Zeyu and Liu, Tong and Fu, Yanwei and Van Gool, Luc and Jiang, Xingqun},
  booktitle={European Conference on Computer Vision},
  year={2024}
}

@inproceedings{guo2020broader,
  title={A broader study of cross-domain few-shot learning},
  author={Guo, Yunhui and Codella, Noel C and Karlinsky, Leonid and Codella, James V and Smith, John R and Saenko, Kate and Rosing, Tajana and Feris, Rogerio},
  booktitle={European Conference on Computer Vision},
  year={2020}
}

@inproceedings{zhou2023revisiting,
  title={Revisiting prototypical network for cross domain few-shot learning},
  author={Zhou, Fei and Wang, Peng and Zhang, Lei and Wei, Wei and Zhang, Yanning},
  booktitle={Proceedings of the IEEE/CVF conference on computer vision and pattern recognition},
  pages={20061--20070},
  year={2023}
}

@article{li2023knowledge,
  title={Knowledge transduction for cross-domain few-shot learning},
  author={Li, Pengfang and Liu, Fang and Jiao, Licheng and Li, Shuo and Li, Lingling and Liu, Xu and Huang, Xinyan},
  journal={Pattern Recognition},
  volume={141},
  pages={109652},
  year={2023},
  publisher={Elsevier}
}

@inproceedings{ghiasi2021simple,
  title={Simple copy-paste is a strong data augmentation method for instance segmentation},
  author={Ghiasi, Golnaz and Cui, Yin and Srinivas, Aravind and Qian, Rui and Lin, Tsung-Yi and Cubuk, Ekin D and Le, Quoc V and Zoph, Barret},
  booktitle={Proceedings of the IEEE/CVF conference on computer vision and pattern recognition},
  pages={2918--2928},
  year={2021}
}

@article{zhang2023diffusionengine,
  title={Diffusionengine: Diffusion model is scalable data engine for object detection},
  author={Zhang, Manlin and Wu, Jie and Ren, Yuxi and Li, Ming and Qin, Jie and Xiao, Xuefeng and Liu, Wei and Wang, Rui and Zheng, Min and Ma, Andy J},
  journal={arXiv preprint arXiv:2309.03893},
  year={2023}
}

@article{bochkovskiy2020yolov4,
  title={Yolov4: Optimal speed and accuracy of object detection},
  author={Bochkovskiy, Alexey and Wang, Chien-Yao and Liao, Hong-Yuan Mark},
  journal={arXiv preprint arXiv:2004.10934},
  year={2020}
}

@article{he2024g,
  title={G-retriever: Retrieval-augmented generation for textual graph understanding and question answering},
  author={He, Xiaoxin and Tian, Yijun and Sun, Yifei and Chawla, Nitesh and Laurent, Thomas and LeCun, Yann and Bresson, Xavier and Hooi, Bryan},
  journal={Advances in Neural Information Processing Systems},
  volume={37},
  pages={132876--132907},
  year={2024}
}

@article{lin2023fine,
  title={Fine-grained late-interaction multi-modal retrieval for retrieval augmented visual question answering},
  author={Lin, Weizhe and Chen, Jinghong and Mei, Jingbiao and Coca, Alexandru and Byrne, Bill},
  journal={Advances in Neural Information Processing Systems},
  volume={36},
  pages={22820--22840},
  year={2023}
}

@inproceedings{ramos2023smallcap,
  title={Smallcap: lightweight image captioning prompted with retrieval augmentation},
  author={Ramos, Rita and Martins, Bruno and Elliott, Desmond and Kementchedjhieva, Yova},
  booktitle={Proceedings of the IEEE/CVF Conference on Computer Vision and Pattern Recognition},
  pages={2840--2849},
  year={2023}
}

@inproceedings{li2024evcap,
  title={Evcap: Retrieval-augmented image captioning with external visual-name memory for open-world comprehension},
  author={Li, Jiaxuan and Vo, Duc Minh and Sugimoto, Akihiro and Nakayama, Hideki},
  booktitle={Proceedings of the IEEE/CVF Conference on Computer Vision and Pattern Recognition},
  pages={13733--13742},
  year={2024}
}

@inproceedings{huang2017arbitrary,  
title={Arbitrary style transfer in real-time with adaptive instance normalization},   author={Huang, Xun and Belongie, Serge},   
booktitle={ICCV},   
year={2017} 
}

@inproceedings{lin2014microsoft,
  title={Microsoft coco: Common objects in context},
  author={Lin, Tsung-Yi and Maire, Michael and Belongie, Serge and Hays, James and Perona, Pietro and Ramanan, Deva and Doll{\'a}r, Piotr and Zitnick, C Lawrence},
  booktitle={ECCV},
  year={2014},
  organization={Springer}
}

@article{lewis2020retrieval,
  title={Retrieval-augmented generation for knowledge-intensive nlp tasks},
  author={Lewis, Patrick and Perez, Ethan and Piktus, Aleksandra and Petroni, Fabio and Karpukhin, Vladimir and Goyal, Naman and K{\"u}ttler, Heinrich and Lewis, Mike and Yih, Wen-tau and Rockt{\"a}schel, Tim and others},
  journal={Advances in neural information processing systems},
  year={2020}
}

@article{zheng2025retrieval,
  title={Retrieval Augmented Generation and Understanding in Vision: A Survey and New Outlook},
  author={Zheng, Xu and Weng, Ziqiao and Lyu, Yuanhuiyi and Jiang, Lutao and Xue, Haiwei and Ren, Bin and Paudel, Danda and Sebe, Nicu and Van Gool, Luc and Hu, Xuming},
  journal={arXiv preprint arXiv:2503.18016},
  year={2025}
}

@article{zou2024closer,
  title={A Closer Look at the CLS Token for Cross-Domain Few-Shot Learning},
  author={Zou, Yixiong and Yi, Shuai and Li, Yuhua and Li, Ruixuan},
  journal={Advances in Neural Information Processing Systems},
  volume={37},
  pages={85523--85545},
  year={2024}
}

@inproceedings{zhang2022free,
  title={Free-lunch for cross-domain few-shot learning: Style-aware episodic training with robust contrastive learning},
  author={Zhang, Ji and Song, Jingkuan and Gao, Lianli and Shen, Hengtao},
  booktitle={Proceedings of the 30th ACM international conference on multimedia},
  pages={2586--2594},
  year={2022}
}

@inproceedings{fu2023styleadv,
  title={StyleAdv: Meta Style Adversarial Training for Cross-Domain Few-Shot Learning},
  author={Fu, Yuqian and Xie, Yu and Fu, Yanwei and Jiang, Yu-Gang},
  booktitle={CVPR},
  year={2023}
}

@inproceedings{zhuo2022tgdm,
  title={TGDM: Target Guided Dynamic Mixup for Cross-Domain Few-Shot Learning},
  author={Zhuo, Linhai and Fu, Yuqian and Chen, Jingjing and Cao, Yixin and Jiang, Yu-Gang},
  booktitle={ACM MM},
  year={2022}
}

@inproceedings{fu2021meta,
  title={Meta-FDMixup: Cross-Domain Few-Shot Learning Guided by Labeled Target Data},
  author={Fu, Yuqian and Fu, Yanwei and Jiang, Yu-Gang},
  booktitle={ACM MM},
  year={2021}
}

@inproceedings{tseng2020cross,
  title={Cross-domain few-shot classification via learned feature-wise transformation},
  author={Tseng, Hung-Yu and Lee, Hsin-Ying and Huang, Jia-Bin and Yang, Ming-Hsuan},
  booktitle = {ICLR},
  year={2020}
}

@inproceedings{liang2021boosting,
  title={Boosting the Generalization Capability in Cross-Domain Few-shot Learning via Noise-enhanced Supervised Autoencoder},
  author={Liang, Hanwen and Zhang, Qiong and Dai, Peng and Lu, Juwei},
  booktitle={ICCV},
  year={2021}
}

@inproceedings{pan2025enhance, 
  title={Enhance Then Search: An Augmentation-Search Strategy with Foundation Models for Cross-Domain Few-Shot Object Detection},
  author={Pan, Jiancheng and Liu, Yanxing and He, Xiao and Peng, Long and Li, Jiahao and Sun, Yuze and Huang, Xiaomeng},
  booktitle={CVPRW},
  year={2025}
}

@article{fu2025ntire,
  title={NTIRE 2025 challenge on cross-domain few-shot object detection: Methods and results},
  author={Fu, Yuqian and Qiu, Xingyu and Ren, Bin and Fu, Yanwei and Timofte, Radu and Sebe, Nicu and Yang, Ming-Hsuan and Van Gool, Luc and Zhang, Kaijin and Nong, Qingpeng and others},
  journal={CVPRW},
  year={2025}
}

@article{cheng2021prototype,
  title={Prototype-CNN for few-shot object detection in remote sensing images},
  author={Cheng, Gong and Yan, Bowei and Shi, Peizhen and Li, Ke and Yao, Xiwen and Guo, Lei and Han, Junwei},
  journal={IEEE Transactions on Geoscience and Remote Sensing},
  volume={60},
  pages={1--10},
  year={2021},
  publisher={IEEE}
}

@inproceedings{fan2020few,
  title={Few-shot object detection with attention-RPN and multi-relation detector},
  author={Fan, Qi and Zhuo, Wei and Tang, Chi-Keung and Tai, Yu-Wing},
  booktitle={Proceedings of the IEEE/CVF conference on computer vision and pattern recognition},
  pages={4013--4022},
  year={2020}
}

@inproceedings{lu2023breaking,
  title={Breaking immutable: Information-coupled prototype elaboration for few-shot object detection},
  author={Lu, Xiaonan and Diao, Wenhui and Mao, Yongqiang and Li, Junxi and Wang, Peijin and Sun, Xian and Fu, Kun},
  booktitle={Proceedings of the AAAI Conference on Artificial Intelligence},
  volume={37},
  number={2},
  pages={1844--1852},
  year={2023}
}

@inproceedings{han2023few,
  title={Few-shot object detection via variational feature aggregation},
  author={Han, Jiaming and Ren, Yuqiang and Ding, Jian and Yan, Ke and Xia, Gui-Song},
  booktitle={Proceedings of the AAAI Conference on Artificial Intelligence},
  volume={37},
  number={1},
  pages={755--763},
  year={2023}
}

@article{podell2023sdxl,
  title={Sdxl: Improving latent diffusion models for high-resolution image synthesis},
  author={Podell, Dustin and English, Zion and Lacey, Kyle and Blattmann, Andreas and Dockhorn, Tim and M{\"u}ller, Jonas and Penna, Joe and Rombach, Robin},
  journal={arXiv preprint arXiv:2307.01952},
  year={2023}
}

@misc{fluxfill2024,
    author={Black Forest Labs},
    title={FLUX.Fill},
    year={2024},
    howpublished={\url{https://huggingface.co/black-forest-labs/FLUX.1-Fill-dev}},
}

@misc{FLUX,
    author={Black Forest Labs},
    title={FLUX},
    year={2024},
    howpublished={\url{https://github.com/black-forest-labs/flux}},
}

@misc{fluxredux2024,
    author={Black Forest Labs},
    title={FLUX.Redux},
    year={2024},
    howpublished={\url{https://huggingface.co/black-forest-labs/FLUX.1-Redux-dev}},
}

@inproceedings{zhang2023adding,
  title={Adding conditional control to text-to-image diffusion models},
  author={Zhang, Lvmin and Rao, Anyi and Agrawala, Maneesh},
  booktitle={Proceedings of the IEEE/CVF international conference on computer vision},
  pages={3836--3847},
  year={2023}
}

@inproceedings{suvorov2022resolution,
  title={Resolution-robust large mask inpainting with fourier convolutions},
  author={Suvorov, Roman and Logacheva, Elizaveta and Mashikhin, Anton and Remizova, Anastasia and Ashukha, Arsenii and Silvestrov, Aleksei and Kong, Naejin and Goka, Harshith and Park, Kiwoong and Lempitsky, Victor},
  booktitle={Proceedings of the IEEE/CVF winter conference on applications of computer vision},
  pages={2149--2159},
  year={2022}
}

@inproceedings{liu2021swin,
  title={Swin transformer: Hierarchical vision transformer using shifted windows},
  author={Liu, Ze and Lin, Yutong and Cao, Yue and Hu, Han and Wei, Yixuan and Zhang, Zheng and Lin, Stephen and Guo, Baining},
  booktitle={Proceedings of the IEEE/CVF international conference on computer vision},
  pages={10012--10022},
  year={2021}
}

@article{loshchilov2017decoupled,
  title={Decoupled weight decay regularization},
  author={Loshchilov, Ilya and Hutter, Frank},
  journal={arXiv preprint arXiv:1711.05101},
  year={2017}
}

@article{zhuo2024prompt,
  title={Prompt as Free Lunch: Enhancing Diversity in Source-Free Cross-domain Few-shot Learning through Semantic-Guided Prompting},
  author={Zhuo, Linhai and Wang, Zheng and Fu, Yuqian and Qian, Tianwen},
  journal={arXiv preprint},
  year={2024}
}

@article{zhuo2024unified,
  title={Unified view empirical study for large pretrained model on cross-domain few-shot learning},
  author={Zhuo, Linhai and Fu, Yuqian and Chen, Jingjing and Cao, Yixin and Jiang, Yu-Gang},
  journal={ACM TOMM},
  year={2024}
}

@inproceedings{fu2022me,
  title={Me-d2n: Multi-expert domain decompositional network for cross-domain few-shot learning},
  author={Fu, Yuqian and Xie, Yu and Fu, Yanwei and Chen, Jingjing and Jiang, Yu-Gang},
  booktitle={ACM Multimedia},
  year={2022}
}

@article{fu2022wave,
  title={Wave-san: Wavelet based style augmentation network for cross-domain few-shot learning},
  author={Fu, Yuqian and Xie, Yu and Fu, Yanwei and Chen, Jingjing and Jiang, Yu-Gang},
  journal={arXiv preprint},
  year={2022}
}

@inproceedings{cao2024chasing,
  title={Chasing day and night: Towards robust and efficient all-day object detection guided by an event camera},
  author={Cao, Jiahang and Zheng, Xu and Lyu, Yuanhuiyi and Wang, Jiaxu and Xu, Renjing and Wang, Lin},
  booktitle={ICRA},
  year={2024}
}

@inproceedings{liu2023spatio,
  title={Spatio-Temporal Graph Diffusion for Text-Driven Human Motion Generation.},
  author={Liu, Chang and Zhao, Mengyi and Ren, Bin and Liu, Mengyuan and Sebe, Nicu and others},
  booktitle={BMVC},
  year={2023}
}

@inproceedings{luo2023closer,
  title={A closer look at few-shot classification again},
  author={Luo, Xu and Wu, Hao and Zhang, Ji and Gao, Lianli and Xu, Jing and Song, Jingkuan},
  booktitle={ICML},
  year={2023}
}

@article{lu2025musia,
  title={MuSIA: Exploiting multi-source information fusion with abnormal activations for out-of-distribution detection},
  author={Lu, Heng-yang and Guo, Xin and Jiang, Wenyu and Fan, Chenyou and Du, Yuntao and Shao, Zhenhao and Fang, Wei and Wu, Xiaojun},
  journal={Neural Networks},
  year={2025}
}

@article{lu2025enhancing,
  title={Enhancing few-shot out-of-distribution intent detection by reducing attention misallocation},
  author={Lu, Heng-yang and Zhang, Jia-ming and Du, Yuntao and Xia, Chang and Wang, Chongjun and Fang, Wei and Wu, Xiao-jun},
  journal={Neurocomputing},
  year={2025}
}

@article{liu2025ot,
  title={OT-DETECTOR: Delving into Optimal Transport for Zero-shot Out-of-Distribution Detection},
  author={Liu, Yu and Tang, Hao and Zhang, Haiqi and Qin, Jing and Li, Zechao},
  journal={IJCAI},
  year={2025}
}

@article{wang2025rag,
  title={RAG-6DPose: Retrieval-Augmented 6D Pose Estimation via Leveraging CAD as Knowledge Base},
  author={Wang, Kuanning and Fu, Yuqian and Wang, Tianyu and Fu, Yanwei and Liang, Longfei and Jiang, Yu-Gang and Xue, Xiangyang},
  journal={IROS},
  year={2025}
}

@article{liu2024multi,
  title={Multi-modal prototypes for few-shot object detection in remote sensing images},
  author={Liu, Yanxing and Pan, Zongxu and Yang, Jianwei and Zhou, Peiling and Zhang, Bingchen},
  journal={Remote Sensing},
  year={2024}
}

@article{wen2025rohoi, 
title={RoHOI: Robustness Benchmark for Human-Object Interaction Detection}, 
author={Wen, Di and Peng, Kunyu and Yang, Kailun and Chen, Yufan and Liu, Ruiping and Zheng, Junwei and Roitberg, Alina and Paudel, Danda Pani and Van Gool, Luc and Stiefelhagen, Rainer}, 
journal={arXiv preprint arXiv:2507.09111}, year={2025}
}

@article{peng2024mitigating, 
title={Mitigating Label Noise using Prompt-Based Hyperbolic Meta-Learning in Open-Set Domain Generalization}, 
author={Peng, Kunyu and Wen, Di and Saquib, Sarfraz M and Chen, Yufan and Zheng, Junwei and Schneider, David and Yang, Kailun and Wu, Jiamin and Roitberg, Alina and Stiefelhagen, Rainer}, 
journal={arXiv preprint arXiv:2412.18342}, year={2024}}

@article{peng2024advancing, 
title={Advancing open-set domain generalization using evidential bi-level hardest domain scheduler}, 
author={Peng, Kunyu and Wen, Di and Yang, Kailun and Luo, Ao and Chen, Yufan and Fu, Jia and Sarfraz, M Saquib and Roitberg, Alina and Stiefelhagen, Rainer}, 
journal={NeurIPS}, year={2024}
}

@article{pan2025earthsynth,
  title={EarthSynth: Generating Informative Earth Observation with Diffusion Models},
  author={Pan, Jiancheng and Lei, Shiye and Fu, Yuqian and Li, Jiahao and Liu, Yanxing and Sun, Yuze and He, Xiao and Peng, Long and Huang, Xiaomeng and Zhao, Bo},
  journal={arXiv preprint arXiv:2505.12108},
  year={2025}
}

@article{anmultimodality,
  title={Multimodality Helps Few-shot 3D Point Cloud Semantic Segmentation},
  author={An, Zhaochong and Sun, Guolei and Liu, Yun and Li, Runjia and Wu, Min and Cheng, Ming-Ming and Konukoglu, Ender and Belongie, Serge},
  journal={ICLR},
  year={2025}
}

@inproceedings{an2025generalized,
  title={Generalized few-shot 3d point cloud segmentation with vision-language model},
  author={An, Zhaochong and Sun, Guolei and Liu, Yun and Li, Runjia and Han, Junlin and Konukoglu, Ender and Belongie, Serge},
  booktitle={CVPR},
  year={2025}
}

@article{liu2025control,
  title={Control Copy-Paste: Controllable Diffusion-Based Augmentation Method for Remote Sensing Few-Shot Object Detection},
  author={Liu, Yanxing and Pan, Jiancheng and Zhang, Bingchen},
  journal={arXiv preprint arXiv:2507.21816},
  year={2025}
}

@inproceedings{zhou2023seeds,
  title={SeeDS: Semantic separable diffusion synthesizer for zero-shot food detection},
  author={Zhou, Pengfei and Min, Weiqing and Zhang, Yang and Song, Jiajun and Jin, Ying and Jiang, Shuqiang},
  booktitle={ACM MM},
  year={2023}
}

@inproceedings{pan2025locate,
  title={Locate anything on earth: Advancing open-vocabulary object detection for remote sensing community},
  author={Pan, Jiancheng and Liu, Yanxing and Fu, Yuqian and Ma, Muyuan and Li, Jiahao and Paudel, Danda Pani and Van Gool, Luc and Huang, Xiaomeng},
  booktitle={AAAI},
  year={2025}
}


\clearpage
\appendix
\section{Technical Appendices and Supplementary Material}

We first provide the implementation details in Sec.~\ref{sec:implementation_details}. Then, in Sec.~\ref{sec:ablation2}, we present additional ablation studies of our proposed Domain-RAG method. Finally, Sec.~\ref{sec:visualization} includes more visualization results and further analysis.

\subsection{Implementation Details.}
\label{sec:implementation_details}
\subsubsection{More Details on Proposed Method.}

\textbf{Domain-Aware Background Retrieval.} 
Considering the large domain gap, we additionally include the inpainted target support set in the background retrieval pool $\mathcal{D}_{base}$. All the images contained in $\mathcal{D}_{base}$ are encoded with a CLIP vision encoder. For the retrieval process, we $m=100$ and $n=5$. In other words, using the inpainted source image as the query, we select the 100 images with the highest cosine similarity in the CLIP embedding space as semantically aligned candidates. From these, we extract style descriptors using the first four layers of a ResNet-50 and choose the 5 images with the smallest $L_2$ distance to the query, yielding the final set of retrieved backgrounds.

\textbf{Domain-Guided Background Generation.} In this phase, the target image and each retrieved image are processed by Redux~\cite{fluxredux2024}; their embeddings are combined by a weighted sum (1.0 for the target, 0.8 for the retrieval) without any additional textual prompt. The resulting embedding is fed into the FLUX~\cite{FLUX} diffusion model with a guidance scale of 2.5 and 50 sampling steps to synthesize a 1024 × 1024 background image.

\textbf{Foreground-Background Composition.} In the generation stage, we employ the FLUX-Fill~\cite{fluxfill2024} model. Given a target image and its background, we supply FLUX-Fill with the source image, an object-mask that excludes the out-painting bounding box, and a prompt embedding extracted from the second-stage background image via the Redux encoder; no additional textual cue is provided and both weights are kept at 1. Because FLUX-Fill struggles with very small inputs, we introduce an adaptive rescaling strategy. For UODD~\cite{jiang2021underwater}, where bounding boxes are tiny, we preserve the aspect ratio and iteratively upsample the image until its longer side exceeds 2048 pixels, then generate a corresponding upsampled mask. Conversely, in ArTaxOr~\cite{GeirArTaxOr}, some images exceed 4000 × 3000 pixels and would exhaust GPU memory; whenever either edge is larger than 2800 pixels, we downsample by an integer factor and create a matching mask. The (up- or down-)sampled target image, its mask, and the Redux embedding are then passed to FLUX-Fill. We keep the guidance scale at 30.0 but modulate the overall noise strength to suit each target, with 0.8 for FISH~\cite{saleh2020realistic} and DIOR~\cite{li2020object}, 0.9 for ArTaxOr and clipart1k~\cite{inoue2018cross}, 0.3 for NEU-DET~\cite{song2013noise} 0.8 for NWPU VHR-10~\cite{niemeyer2014contextual}, 0.6 for Camouflage~\cite{nguyen2024art} FSOD benchmark, and 0.4 for UODD.

\subsubsection{More Details on Training.}
During training, following Grounding DINO~\cite{liu2024grounding} and ETS~\cite{pan2025enhance}, we apply diverse data augmentations, including Mosaic, MixUp, color jittering, random flipping, multi-scale resizing, and cropping, to improve few-shot generalization of both the base GroundingDINO and our Domain-RAG. 
For evaluation, we follow COCO metrics and disable all augmentations.

\subsection{More Ablation Studies.}
\label{sec:ablation2}
In Sec.~\ref{sec:more-analysis}, we reported the ablation study on each proposed module using NEU-DET under the 1-shot setting as a representative example. To provide a more comprehensive analysis, in Tab.~\ref{tab:ablation_study}, we report the same ablation experiments but conducted on all six CD-FSOD target domains. This includes evaluations of the impact of removing domain-aware background retrieval, domain-guided background generation, and compositional generation stages.

As shown in Tab.~\ref{tab:ablation_study}, 
starting from the baseline GroundingDINO (average 26.3), incorporating our modules makes our full Domain-RAG model attain the highest average mAP (33.6), validating the complementary benefits of domain-aware background retrieval, generation, and compositional synthesis for cross-domain few-shot detection.
More specifically, 1) removing the background retrieval module causes a clear drop in average performance (33.6 to 31.8), demonstrating its essential role in capturing domain-relevant context. 2) Without background generation, the model achieves a slightly lower average (31.1) and notably fails on UODD, indicating less stable generalization. 3) Replacing the compositional synthesis stage with a simple copy-paste strategy further degrades performance to 29.2, confirming the advantage of our generative approach.

\begin{table*}[h!]
\centering
\footnotesize
\vspace{-0.1in}
\caption{Ablation study results under the 1-shot setting on different datasets. }
\vspace{+0.1in}
\label{tab:ablation_study}
\resizebox{0.95\textwidth}{!}{
\begin{tabular}{lccccccc}
\toprule
\textbf{Method} & \textbf{ArTaxOr} & \textbf{Clipart} & \textbf{DIOR} & \textbf{DeepFish} & \textbf{NEU-DET} & \textbf{UODD} & \textbf{Avg.} \\
\midrule
Baseline (GroundingDINO)      & 26.3 & 55.3 & 14.8 & 36.4 & 9.3  & 15.9 & 26.3 \\
w/o Background Retrieval       & 51.7 & 57.3 & 17.0 & 37.2 & 10.9 & 16.4 & 31.8 \\
w/o Background Generation      & 48.4 & 54.6 & 17.2 & 37.9 & 10.2 & 18.0   & 31.1 \\
Copy-Paste as Composition    & 46.6 & 56.2 & 16.7 & 35.0 & 9.1  & 11.7 & 29.2 \\
\textbf{Domain-RAG (Ours)}     & \textbf{57.2} & \textbf{56.1} & \textbf{18.0} & \textbf{38.0} & \textbf{12.1} & \textbf{20.2} & \textbf{33.6} \\
\bottomrule
\vspace{-0.2in}
\end{tabular}}
\end{table*}

\subsubsection{Ablation on Domain-Aware Background Retrieval.}
\noindent\textbf{Ablation on Different Database.}
we set COCO as our database in our main exp setting, while we are also interested in validate our idea on other dataset. Therefore, we conduct additional experiments where we replace COCO with MiniImageNet as the retrieval source. 
As shown in Tab.~\ref{tab:ablation-database}, COCO achieves better performance on most domains, likely due to its richer scene diversity and natural statistics. Performance on DeepFish remains similar across both databases. MiniImageNet is originally designed for classification tasks, and compared to the COCO dataset, it typically contains larger foreground objects and less informative background content. As a result, using MiniImageNet as the retrieval database leads to weaker performance than using COCO. Nevertheless, it still outperforms the baseline, indicating the effectiveness of retrieval-based background generation.

\begin{table}[h]
\centering
\footnotesize
\caption{Ablation on different background retrieval databases for 1-shot CD-FSOD across six datasets.}
\label{tab:ablation-database}
\renewcommand{\arraystretch}{0.9}
\resizebox{0.95\textwidth}{!}{
\begin{tabular}{lccccccc}
\toprule
\textbf{Database} & \textbf{ArTaxOr} & \textbf{Clipart} & \textbf{DIOR} & \textbf{DeepFish} & \textbf{NEU-DET} & \textbf{UODD} & \textbf{Avg.} \\
\midrule
Baseline        & 26.3 & 55.3 & 14.8 & 36.4 & 9.3  & 15.9 & 26.3 \\
MiniImagenet    & 55.6 & 53.2 & 15.6 & \textbf{38.0} & \textbf{14.0} & 16.2 & 32.1 \\
\textbf{COCO (Ours)}     & \textbf{57.2} & \textbf{56.1} & \textbf{18.0} & \textbf{38.0} & 12.1 & \textbf{20.2} & \textbf{33.6} \\
\bottomrule
\end{tabular}
}
\end{table}

\noindent\textbf{Ablation on Retrieval Strategy.} As stated in Sec.~\ref{method}, we use both of the CLIP-semantic and ResNet-style to retrieve the background images from $\mathcal{D}_{base}$; thus, we compare our method with only CLIP-semantic, ResNet-style. Compared to using only CLIP features or only style features, our method achieves higher average performance, as shown in Tab.~\ref{tab:retrieval_strategy_ablation}. This demonstrates the effectiveness of combining both CLIP and style features in our framework.

\begin{table}[h]
\centering
\caption{Ablation on retrieval strategy under the 1-shot setting. We compare CLIP-based, ResNet-style, and our combined retrieval strategy. Results are reported on six CD-FSOD benchmarks.}
\label{tab:retrieval_strategy_ablation}
\footnotesize
\resizebox{0.95\textwidth}{!}{
\begin{tabular}{lccccccc}
\toprule
\textbf{Retrieval Strategy} & \textbf{ArTaxOr} & \textbf{Clipart1k} & \textbf{DIOR} & \textbf{DeepFish} & \textbf{NEU-DET} & \textbf{UODD} & \textbf{Avg.} \\
\midrule
Baseline & 26.3 & 55.3 & 14.8 & 36.4 & 9.3 & 15.9 & 26.3 \\
CLIP Only        & 50.0 & \textbf{56.2} & 16.3 & 36.0 & \textbf{13.0} & 17.7 & 31.5 \\
Style Only       & 48.8 & 54.5 & 16.7 & 35.5 & 12.1 & 16.5 & 30.7 \\
\textbf{CLIP + Style (Ours)} & \textbf{57.2} & 56.1 & \textbf{18.0} & \textbf{38.0} & 12.1 & \textbf{20.2} & \textbf{33.6} \\
\bottomrule
\end{tabular}}
\end{table}

\noindent\textbf{Ablation on Number of Retrieved Images.}
We set $m=5$ in our main results. Here, we also conduct different choices of $m$, such as $m=1$, $m=3$, and $m=10$, to better investigate our methods.
As summarized in Tab.~\ref{tab:retrieved_image_ablation}, 
we observe that retrieving $m=5$ images yields the best performance under the 1-shot setting. Smaller values ($m=1$ or $m=3$) provide limited diversity and result in weaker generalization. In contrast, setting $m=10$ often leads to a performance drop, suggesting that more retrieved samples do not always bring further gains.
This trend is largely due to our data generation process, which composes retrieved foregrounds with background scenes. Too few retrieved images limit visual variation, while too many increase the chance of unrealistic compositions (e.g., a sofa on a grass field). These unnatural contexts may confuse the model and reduce its ability to align with real-world test distributions.

\begin{table}[h]
\centering
\footnotesize
\caption{Ablation on the number of retrieved images ($m$) under the 1-shot setting. Results are reported on six CD-FSOD benchmarks.}
\label{tab:retrieved_image_ablation}
\resizebox{0.95\textwidth}{!}{
\begin{tabular}{lccccccc}
\toprule
\textbf{Retrieved Images} & \textbf{ArTaxOr} & \textbf{Clipart1k} & \textbf{DIOR} & \textbf{DeepFish} & \textbf{NEU-DET} & \textbf{UODD} & \textbf{Avg.} \\
\midrule
Baseline     & 26.3 & 55.3 & 14.8 & 36.4 & 9.3  & 15.9 & 26.3 \\
$m = 1$      & 43.2 & 54.3 & 16.6 & 37.1 & 9.4  & 14.7 & 29.2 \\
$m = 3$      & 49.8 & 54.5 & 16.3 & 36.7 & 12.7 & 18.1 & 31.4 \\
\textbf{$m = 5$ (Ours)} & \textbf{57.2} & 56.1 & \textbf{18.0} & 38.0 & 12.1 & \textbf{20.2} & \textbf{33.6} \\
$m = 10$     & 49.6 & \textbf{56.8} & 16.6 & \textbf{39.3} & \textbf{13.2} & 17.3 & 32.1 \\
\bottomrule
\vspace{-0.2in}
\end{tabular}}
\end{table}

\subsubsection{Ablation on Domain-Guided Background Generation.}
\noindent\textbf{Ablation on with/without Initial Background Guidance.} 
For the Ablation on with/without initial background guidance, we keep all other components of the framework unchanged. During background generation, we do not use the initial background. Instead, we only use the retrieval image to obtain the prompt embedding via Redux, and then feed 0.8 × the retrieval image’s prompt embedding into the Flux model to generate the background.
As shown in Tab.~\ref{tab:ablation-bg-guidance}, incorporating initial background guidance consistently improves performance under the 1-shot setting across six CD-FSOD datasets. Removing this guidance leads to a noticeable drop in average accuracy, especially on challenging domains like UODD. This demonstrates that initial background information plays an important role in stabilizing and enhancing background generation quality.

\begin{table}[h]
\centering
\footnotesize
\caption{Ablation on initial background guidance for 1-shot CD-FSOD across six datasets.}
\label{tab:ablation-bg-guidance}
\resizebox{0.95\textwidth}{!}{
\begin{tabular}{lccccccc}
\toprule
\textbf{Method} & \textbf{ArTaxOr} & \textbf{Clipart1k} & \textbf{DIOR} & \textbf{DeepFish} & \textbf{NEU-DET} & \textbf{UODD} & \textbf{Avg.} \\
\midrule
Baseline  & 26.3 & 55.3 & 14.8 & 36.4 & 9.3  & 15.9 & 26.3 \\
W/o Bg Guidance & 49.8 & 56.0 & 16.3 & \textbf{38.0} & \textbf{12.7} & 15.4 & 31.4 \\
\textbf{With Bg Guidance (Ours)} & \textbf{57.2} & \textbf{56.1} & \textbf{18.0} & \textbf{38.0} & 12.1 & \textbf{20.2} & \textbf{33.6} \\
\bottomrule
\end{tabular}}
\end{table}

\noindent\textbf{Ablation on Text-to-Image Backbones.} In this section, we conduct ablation experiments on the background generation module. In the proposed framework, we use the Redux module to fuse features from the retrieval image and the inpainted image, and convert them into a prompt embedding.
For the ablation setting, we replace Redux with InstructBLIP and use Diffusion XL to generate backgrounds. Specifically, we use InstructBLIP to extract captions for both the retrieval image and the inpainted image. The resulting captions are then concatenated in the form of "$<caption_1>. <caption_2>$". Since Diffusion XL uses a CLIP text encoder that cannot handle excessively long texts, we constrain InstructBLIP to generate captions of no more than 20 words, and truncate any prompt exceeding 40 words before feeding it into the model.  In this ablation, the background generation strategy is only applied to the domain-guided background generation, while all other components of the pipeline remain unchanged.
As shown in Tab.~\ref{tab:ablation-text2img-backbones}, generating richer backgrounds contributes positively to the final generation quality. Our adopted Flux + Redux approach provides more effective supervision, guiding the generation process to produce data that better aligns with the target domain. 

\begin{table}[h]
\centering
\footnotesize
\caption{Ablation on text-to-image backbones for 1-shot CD-FSOD across six datasets.}
\label{tab:ablation-text2img-backbones}
\resizebox{0.95\textwidth}{!}{
\begin{tabular}{lccccccc}
\toprule
\textbf{Method} & \textbf{ArTaxOr} & \textbf{Clipart} & \textbf{DIOR} & \textbf{DeepFish} & \textbf{NEU-DET} & \textbf{UODD} & \textbf{Avg.} \\
\midrule
Baseline  & 26.3 & 55.3 & 14.8 & 36.4 & 9.3 & 15.9 & 26.3 \\
InstructBLIP + Diffusion XL & 49.0 & 54.2 & 15.8 & \textbf{38.6} & 11.8 & 14.3 & 30.6 \\
\textbf{Redux + Flux (Ours)} & \textbf{57.2} & \textbf{56.1} & \textbf{18.0} & 38.0 & \textbf{12.1} & \textbf{20.2} & \textbf{33.6} \\
\bottomrule
\end{tabular}}
\end{table}

\subsubsection{Ablation on Foreground-Background Composition.}
\noindent\textbf{Ablation on Text-to-Image Backbones.}
In our framework, we utilize the semantic information provided by Redux in the third stage to guide image generation. In the ablation experiments for this stage, we replace the Redux module with InstructBLIP, and substitute the Flux-Fill model with the Diffusion-XL-Inpaint model to generate new backgrounds.
Specifically, for each background retrieved during the background retrieval stage, we use InstructBLIP to extract a caption prompt describing the image. After obtaining the prompt, we feed the Target image, the corresponding mask, and the caption (as an instruction to modify the background) into the Diffusion-XL-Inpaint model to synthesize a new background.
As shown in Tab.~\ref{tab:text2image_ablation}, compared to the baseline using InstructBLIP and Diffusion-XL-Inpaint, our method (Redux + Flux) achieves consistently better performance, especially on ArTaxOr and UODD, demonstrating the effectiveness of Redux guidance and Flux in generating semantically coherent and diverse images.

\begin{table}[h]
\centering
\footnotesize
\caption{Ablation on text-to-image backbones under the 1-shot setting. Results are reported on six CD-FSOD benchmarks.}
\label{tab:text2image_ablation}
\resizebox{0.95\textwidth}{!}{
\begin{tabular}{lccccccc}
\toprule
\textbf{Backbone} & \textbf{ArTaxOr} & \textbf{Clipart1k} & \textbf{DIOR} & \textbf{DeepFish} & \textbf{NEU-DET} & \textbf{UODD} & \textbf{Avg.} \\
\midrule
Baseline  & 26.3 & 55.3 & 14.8 & 36.4 & 9.3 & 15.9 & 26.3 \\
Text2Image          & 44.4 & 54.5 & 14.7 & 35.1 & 11.9 & 15.6 & 29.4 \\
\textbf{Redux + Flux (Ours)} & \textbf{57.2} & \textbf{56.1} & \textbf{18.0} & \textbf{38.0} & \textbf{12.1} & \textbf{20.2} & \textbf{33.6} \\
\bottomrule
\vspace{-0.2in}
\end{tabular}}
\end{table}

\subsection{More Visualization and Analysis.}
\label{sec:visualization}

\subsubsection{More Analysis on Generated Image Quantity.}
We evaluate the quality of generated images using CLIP-I similarity and Fréchet Inception Distance (FID). CLIP-I measures the average cosine similarity between the target image and generated samples using the CLIP image encoder, while FID assesses visual fidelity based on InceptionV3 features. Due to its reliance on distribution statistics, FID is not computed for DeepFish in the 1-shot setting, where only one target image is available. Higher CLIP-I and lower FID indicate better semantic alignment and image realism, respectively.

\begin{table*}[h]
\centering
\scriptsize
\vspace{-0.1in}
\caption{Evaluation of different augmentation methods on the CD-FSOD 1-shot benchmark using CLIP-I and FID.}
\vspace{+0.1in}
\label{tab:cd_fsod_clip_fid}
\resizebox{\textwidth}{!}{
\begin{tabular}{l|cc|cc|cc|cc|cc|cc}
\toprule
\multirow{2}{*}{\textbf{Method}} 
& \multicolumn{2}{c|}{\textbf{ArTAXOr}} 
& \multicolumn{2}{c|}{\textbf{Clipart}} 
& \multicolumn{2}{c|}{\textbf{DIOR}} 
& \multicolumn{2}{c|}{\textbf{DeepFish}} 
& \multicolumn{2}{c|}{\textbf{NEU-DET}} 
& \multicolumn{2}{c}{\textbf{UODD}} \\
& CLIP-I & FID 
& CLIP-I & FID 
& CLIP-I & FID  
& CLIP-I & FID 
& CLIP-I & FID 
& CLIP-I & FID \\
\midrule
GroundingDINO & -& -& -& -& -& -& -& -& -& -& -& -\\
Copy-Paste     & 62.9& 321.4& 60.1& 312.8& 52.8& 353.6& 60.9& -& 49.7& 476.2& 50.4& 396.2\\
Foreground Aug & 88.1& 165.0& 89.3& 107.7& 95.1& 131.1& 93.7& -& 89.3& 129.3& 90.9& 27.0\\
Background Aug & 89.7& 78.7& 85.0& 128.0& 83.7& 265.4& 68.9& -& 72.8& 453.7& 78.1& 287.8\\
\textbf{Domain-RAG (Ours)} & 92.6& 70.5& 88.5& 117.6& 79.8& 288.8& 77.1& -& 93.3& 127.4& 79.3& 289.6\\
\bottomrule
\end{tabular}}
\end{table*}

\begin{figure}[h]
    \centering
    \includegraphics[width=0.9\linewidth]{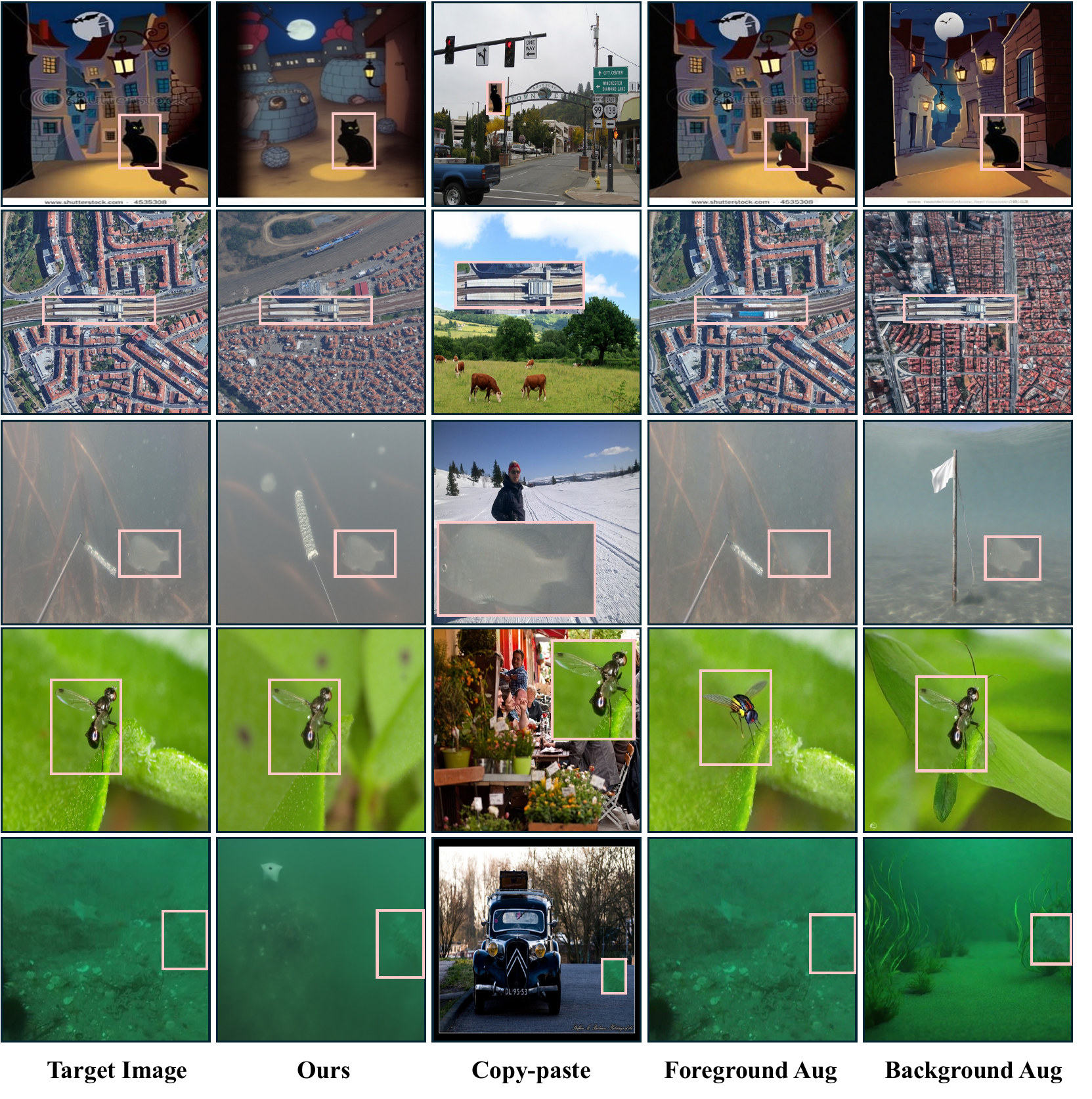}
    \caption{Visualization comparison between DomainRAG and other augmentation methods.}
    \label{fig:Comparison-with-Other-Augmentation-Methods}
\end{figure}

As shown in Tab.~\ref{tab:cd_fsod_clip_fid} and visualized in Fig.~\ref{fig:Comparison-with-Other-Augmentation-Methods}, the "Copy-Paste" method performs poorly on both CLIP-I and FID due to its use of randomly selected COCO backgrounds, resulting in a distribution mismatch with the target domain. "Foreground Aug" yields very high CLIP-I and low FID, as only small local regions are modified, making the overall image similar to the original image. In contrast, "Background Aug" alters large portions of the image, leading to lower CLIP-I and higher FID. Our Domain-RAG achieves a balanced trade-off, maintaining domain relevance while introducing sufficient visual diversity.

In addition, we would like to argue that though CLIP and InceptionV3 are widely used for image similarity evaluation, their general-purpose nature can lead to unreliable assessments in cross-domain settings. Our goal is not to produce images that are distributionally identical to the originals, but to enrich semantic diversity while preserving domain-specific features. Therefore, higher CLIP-I or lower FID does not necessarily indicate better generation quality in our case. Instead, the quantitative results demonstrate the effectiveness of our approach.

\subsubsection{Visualization Results from Each Stage.}

To better illustrate the effectiveness and progression of our data generation pipeline, we provide visualizations of intermediate outputs from each stage, as shown in Fig.\ref{fig:abla}.
From left to right, Fig.\ref{fig:abla} presents:  
(1) the target query image,  
(2) the retrieved support images from the domain-aware background retrieval stage,  
(3) the generated background images from the domain-guided background generation stage,
(4) the final synthesized image from the foreground-background composition stage, which is ultimately used for training.
This progressive visualization highlights how each stage contributes to generating diverse and semantically meaningful training samples that align with the target domain.

\begin{figure}[h]
    \vskip -0.1in
    \centering
    \centerline{\includegraphics[width=\columnwidth]{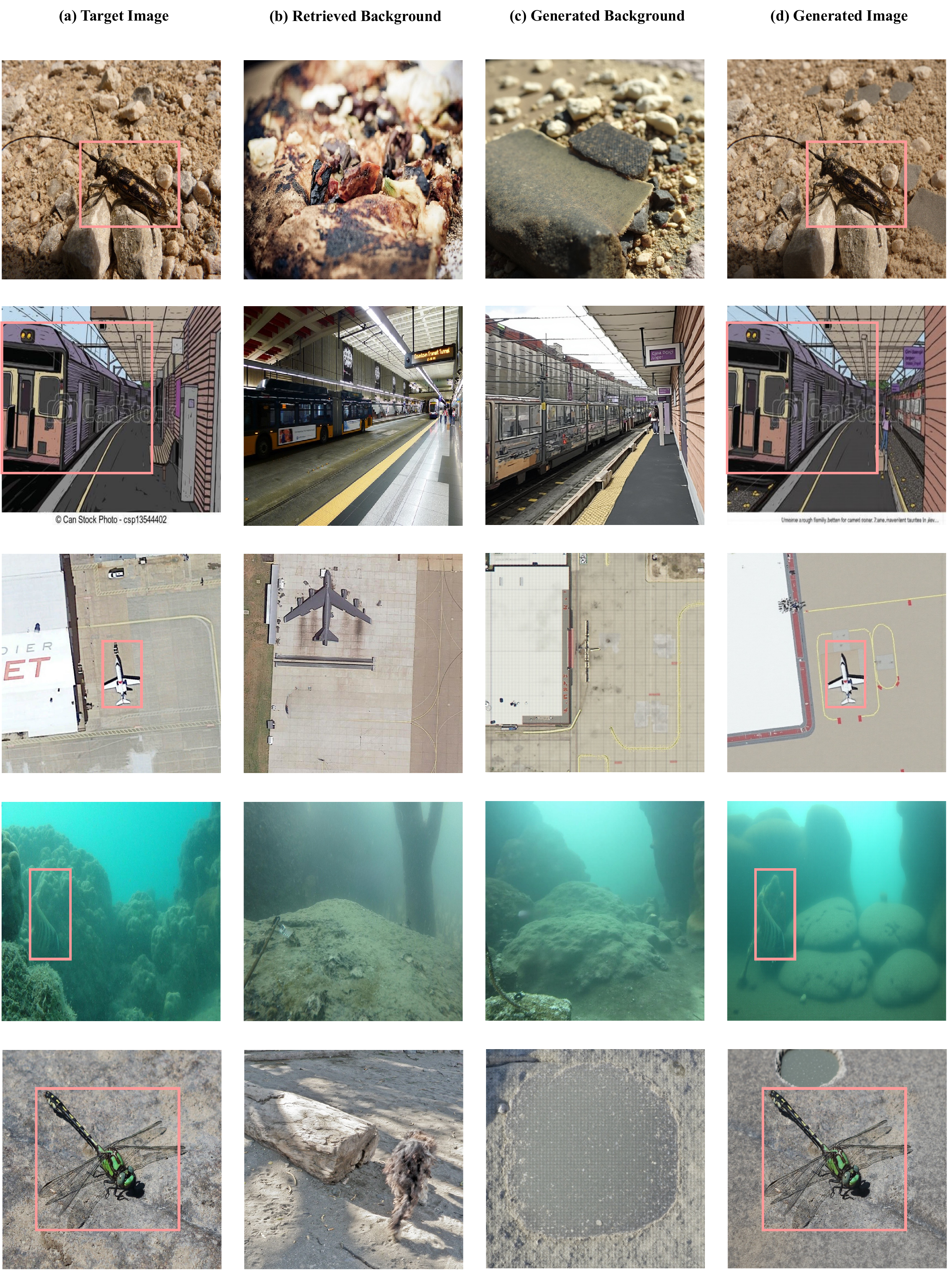}} 
    \caption{Visualization of our data generation pipeline. From left to right: (1) the target query image, (2) retrieved backgrounds, (3) generated new backgrounds, and (4) the final synthesized image used for further model finetuning.} 
    \label{fig:abla}
    \vskip -0.1in
\end{figure}

\subsubsection{More Visualization Results of our Domain-RAG.}

\begin{figure}[h]
\centering
    \centerline{\includegraphics[width=\columnwidth]{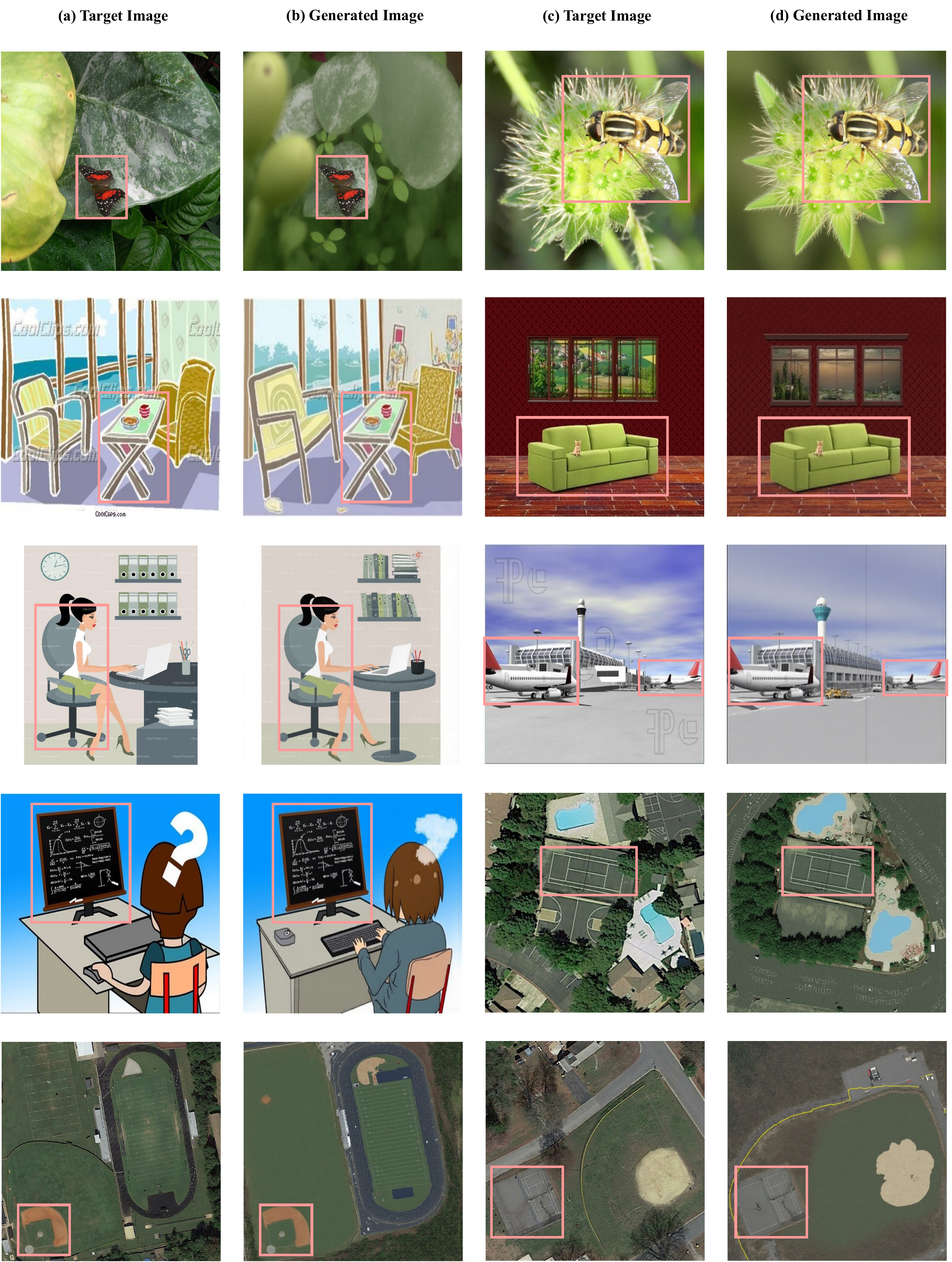}}
    \caption{More visualization results of our Domain-RAG module across different domains. Each row shows two pairs of query images and their corresponding generated backgrounds. Our method demonstrates the ability to generate semantically aligned and domain-aware backgrounds across diverse visual domains such as artistic, aerial, underwater, and industrial scenes.}
        \vskip -0.1in
\label{fig:domain-rag-vis-full}
\end{figure}

Fig.\ref{fig:domain-rag-vis-full} presents additional qualitative results of our Domain-RAG module across multiple domains. Each row shows two examples, arranged from left to right as: (a) target image, (b) generated image, (c) target image, and (d) generated image.  The results demonstrate that our method generates visually consistent and domain-aware backgrounds across diverse visual styles, including artistic, aerial, underwater, and industrial scenes.

\subsubsection{More Analysis on Limitations and Future Work.}
As stated in Sec.~\ref{sec:more-analysis}, our model exhibits foreground information leakage. We attribute this issue to the limited generation quality of Simple LaMa Inpaint—for target images with large foregrounds, the inpainting results are often suboptimal, frequently showing blurriness or patch artifacts after foreground removal. The pretrained Redux module typically struggles to handle these artifacts effectively, often preserving them in subsequent generation steps, which in turn degrades the overall generation quality. Enhancing the capability of the Redux module to better support such cases and mitigate foreground leakage remains an important direction for future work. In addition, we currently lack an effective filtering mechanism. Mostly commonly used and general-purpose vision backbones, e.g., CLIP, fail to deliver ideal filtering results in our cross-domain scenarios.
To address this limitation, future work could explore the integration of more powerful vision-language models to enable more precise and background-aware filtering strategies. 

\end{document}